\documentclass[lettersize,journal]{IEEEtran}
\usepackage{amsmath,amsfonts}
\usepackage{algorithmic}
\usepackage{algorithm}
\usepackage{array}
\usepackage[caption=false,font=normalsize,labelfont=sf,textfont=sf]{subfig}
\usepackage{textcomp}
\usepackage{stfloats}
\usepackage{url}
\usepackage{verbatim}
\usepackage{graphicx}
\usepackage{cite}
\usepackage{tikz}
\usetikzlibrary{arrows,shapes,positioning,shadows,trees}
\usepackage{forest}
\usepackage{tikz-qtree}
\usetikzlibrary{arrows.meta}
\usepackage[export]{adjustbox}
\usepackage{float} 
\usepackage{booktabs}
\usepackage{multirow}
\usepackage{tabularx}
\usepackage{diagbox}
\usepackage{xcolor}

\usepackage{amsmath}

\DeclareMathOperator*{\argmin}{arg\,min}
\newcommand{\etal}{{et al}. }
\newcommand{\etals}{{et al}.'s}

\usepackage{amsfonts}

\begin{document}

\title{Evaluation in Neural Style Transfer: A Review}

\author{Eleftherios Ioannou and Steve Maddock \\
The University of Sheffield, UK}



\maketitle

\begin{abstract}
The field of Neural Style Transfer (NST) has witnessed remarkable progress in the past few years, with approaches being able to synthesize artistic and photorealistic images and videos of exceptional quality. To evaluate such results, a diverse landscape of evaluation methods and metrics is used, including authors' opinions based on side-by-side comparisons, human evaluation studies that quantify the subjective judgements of participants, and a multitude of quantitative computational metrics which objectively assess the different aspects of an algorithm's performance. However, there is no consensus regarding the most suitable and effective evaluation procedure that can guarantee the reliability of the results.
In this review, we provide an in-depth analysis of existing evaluation techniques, identify the inconsistencies and limitations of current evaluation methods, and give recommendations for standardized evaluation practices. We believe that the development of a robust evaluation framework will not only enable more meaningful and fairer comparisons among NST methods but will also enhance the comprehension and interpretation of research findings in the field. 
\end{abstract}


\section{Introduction}

\label{sec:Introduction}

Neural Style Transfer (NST) refers to the class of methods that attempt to synthesize artistically stylized visual media. For images, where it is most predominantly applied, NST 
is the process of transferring the style of an artistic image onto an ordinary photograph, a process pioneered in the seminal work of Gatys~\etal \cite{gatys2015neural}.
Successive visual synthesis models have demonstrated remarkable improvements, and their capability to generate impressive outputs have attracted the attention of both academia and industry. Besides the significant advancements to the quality of the generated results, style transfer has also been extended to work for a range of media: images, videos, 3D meshes, point clouds, and radiance fields. 

To assess the quality of the results of style transfer methods, there exists an array of diverse evaluation techniques. Visual side-by-side comparisons are used to evaluate results qualitatively; user surveys are employed to quantitatively gauge the performance of the proposed models from collected subjective responses; and computational metrics attempt to quantify content retainment, style resemblance and overall efficiency. 
Despite the availability of a plethora of evaluation methodologies, the evaluation process differs amongst the NST algorithms.  Depending on the medium, but also on the particular advancement or contribution a method aims to achieve, each style transfer approach resorts to different evaluation procedures. As a result, there are no established benchmarks or evaluation protocols.

The challenge in establishing a universally accepted evaluation protocol is, in part, rooted in the definition of NST. Artistic NST algorithms seek to generate outputs that resemble the reference artwork while retaining the content of the image that is stylized -- by definition, the most significant aspects of the evaluation of NST results are \textit{content preservation} and \textit{style performance}. However, there is no agreement as to where the line between content fidelity and stylization intensity should be optimally drawn. This observation magnifies the difficulty of the task, which arguably encapsulates a degree of subjectivity. The subjectivity involved in the assessment of the outputs of NST methods, the particular intentions and goals of each approach, and the perspective from which each method attempts to tackle the problem of stylization, render evaluation a complicated and demanding activity.

The existing evaluation approaches in NST research are typically classified into two categories: Qualitative and Quantitative \cite{jing2019neural}. \emph{Qualitative Evaluation} relies upon the subjective judgements of the corresponding authors. \emph{Quantitative Evaluation} includes numerous computational metrics 
to derive exact numerical estimations of particular aspects of the performance of the evaluated approach, as well as quantitative data from user studies that can be statistically interpreted. 

In current literature, however, user survey results are often excluded from the Qualitative or Quantitative evaluation sections of the studies. The majority of the approaches provide the human evaluation studies' details and results in distinct sections, with minor exceptions \cite{liu2021learning,sanakoyeu2018style}. Notwithstanding, the user studies conducted in NST only gather quantitative data: the participants are either asked to choose their favorite stylization or rank a set of images, without any textual feedback or qualitative observations being collected. Consequently, it is justifiable to classify the data collected from the NST user studies as quantitative. While collecting qualitative data from user surveys could be a useful and insightful practice, in their current state, user surveys are solely utilized for the collection and analysis of quantitative data. 
The results of a user study can be depicted numerically and can allow for quantitative assessments; yet, other vital concerns arise, such as the necessity to estimate and explicitly indicate the statistical significance of those results. Reproducibility and repeatability are also crucial factors that should be taken into account to ensure the reliability of the reported results, similar to the results calculated using computational metrics.

\begin{figure}[htb]
\centering
\resizebox{\linewidth}{!}{%
\begin{tikzpicture}[
    level 1/.style = {black, sibling distance = 1.2cm, level distance = 2cm},
  level 2/.style = {black, level distance = 1.9cm, sibling distance = 1.2cm},
  edge from parent fork down,
    every node/.append style = {draw}]
\node { Evaluation in NST}
  child {node[style=blue] {\begin{tabular}[c]{@{}l@{}} Qualitative\\ Evaluation\end{tabular}}
    child { 
        child {node[style=black] {\begin{tabular}[c]{@{}l@{}}Authors' \\ Subjective\\ Judgements\end{tabular}}}
    }
  }
  child [missing] {} 
  child [missing] {}
  child [missing] {}
  child [missing] {}
  child {node[style=black] {\begin{tabular}[c]{@{}l@{}}Quantitative\\ Evaluation\end{tabular}}
    child {node[style=blue] {\begin{tabular}[c]{@{}l@{}} Human \\ Evaluation\\ Studies\end{tabular}}}
    child [missing] {}
    child [missing] {}
    child [missing] {}
    child [missing] {}
    child {node[style=blue] {\begin{tabular}[c]{@{}l@{}}Computational\\ Metrics\end{tabular}}
        child {node[style=black] {\begin{tabular}[c]{@{}l@{}}Perceptual\\ Metrics\end{tabular}}}
        child [missing] {}
        child {node[style=black] {\begin{tabular}[c]{@{}l@{}}Stylization\\ Performance \\ Metrics\end{tabular}}}
        child [missing] {}
        child {node[style=black] {\begin{tabular}[c]{@{}l@{}}Video\\ Metrics\end{tabular}}}
        child [missing] {}
        child {node[style=black] {\begin{tabular}[c]{@{}l@{}}Efficiency\\Metrics\end{tabular}}}
    }
  };
\end{tikzpicture}%
}
\caption{Evaluation in NST: Qualitative Evaluation and Quantitative Evaluation. To align with current literature, we use three main categories for evaluation, highlighted in blue.}
\label{fig:evaluation-general-categories}
\end{figure}
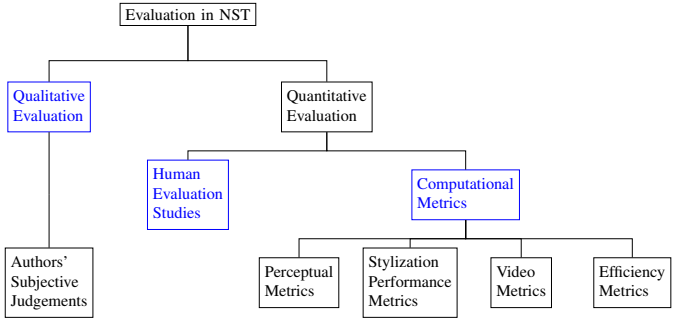

An overview of the evaluation landscape in NST is given in Figure~\ref{fig:evaluation-general-categories}. Human evaluation studies and computational metrics are included in the Quantitative Evaluation category. However, to be consistent with current literature, we structure our review into three evaluation areas (highlighted in blue): \emph{Qualitative Evaluation}, \emph{Human Evaluation Studies}, and \emph{Quantitative Evaluation Metrics} (Figure~\ref{fig:evaluation-taxonomy} gives an in-depth illustration of these areas). All approaches provide a qualitative evaluation to some extent -- visual side-by-side results are presented accompanied by subjective assessments of the authors. Whilst user studies are quite commonplace, not every method agrees on the structure of the user study that is conducted. Also, virtually all the methods quantitatively gauge the performance of their system selectively, drawing from a pool of different computational metrics. This results in a variety of different metrics being present in each of the methods' evaluation sections, with no shared list of metrics being agreed upon.

Arguably, evaluation is an essential aspect of NST research and one that is catalytic to the success and progress of the field, as it can provide a solid understanding of the proposed approaches and a bedrock for novel research. This work aims to provide a comprehensive review of the current evaluation methods and identify the commonalities and differences between the state-of-the-art approaches. An extensive, in-depth survey of the evaluation techniques in NST would provide useful insights and potential solutions for the establishment of robust and reliable evaluation methodologies. 

Previous review papers \cite{jing2019neural,singh2021neural} have analyzed the techniques and applicability of NST, but have not focused on evaluation methodologies.
To our knowledge, this is the first work that directly considers the problem of evaluation in NST, reviewing all the approaches to evaluation and their limitations. 
As part of the review, we aim to highlight the most important evaluation techniques and propose suggestions for the design of a standardized evaluation procedure. 

The rest of the paper is organised as follows. \textbf{Section~\ref{sec:nst-methods}} provides an overview of the NST literature, distinguishing the different approaches by the visual medium they are applied to. \textbf{Section~\ref{sec:current-evaluation}} examines the evaluation techniques and metrics. \textbf{Section~\ref{sec:analysis}} presents a review of the evaluation data and the qualitative and quantitative evaluation approaches in the current literature. It also provides an analysis of those methods, complemented with an array of recommendations. 


\section{Neural Style Transfer Methods}
\label{sec:nst-methods}

Jing~\etal \cite{jing2019neural} categorize NST algorithms into Image-Optimization-Based Online Neural methods (which optimize a generative model online) and Model-Optimization-Based Offline Neural methods (which optimize a generative model offline, then produce a stylization with a single forward pass). The lower tiers in their hierarchical taxonomy split the methods into Non-Photorealistic and Photorealistic branches with the ending nodes further classifying the methods into Image and Video. 
For this paper, we follow a similar taxonomy by distinguishing the reviewed NST methods by the medium they are applied to: images and videos.

We specifically concentrate on models that, once trained, are capable of synthesizing a stylized output given an input content image/video (and potentially a reference style image depending on the network's capability to reproduce multiple or arbitrary styles). We do not include text-to-image techniques, such as those utilizing \textit{CLIP} \cite{radford2021learning}. Our review of evaluation techniques is based on the analysis of a single output. Consequently, we do not assess different aspects of stylization, such as interpolation between various styles. 
Moreover, although NST is also applied to 3D data, here we mainly focus on the evaluation techniques used by NST methods applied to images (Tables~\ref{tbl1}-\ref{tbl6}) and/or videos (Table~\ref{tbl7}).


\begin{table}[htb]
\setlength\extrarowheight{1pt}
\centering
    \caption{Online Image Optimization Methods. \textit{`I'} under User Study denotes `Informal'.} \label{tbl1}

    \resizebox{1\linewidth}{!}{%

    \begin{tabular}{ p{2em}  p{11em} | p{1em} | p{1em}| p{0.2em} | p{1em} | p{1em} | p{10em}}

    \toprule
        \multicolumn{1}{l}{\textbf{Year}} & \multicolumn{1}{l}{\textbf{Method}} & \multicolumn{3}{l}{\textbf{Mode}} & \multicolumn{3}{l}{\textbf{Evaluation}}   \\ 
    \midrule 
            \multicolumn{1}{c}{} & {} & {\rotatebox{90}{\textbf{Artistic}}} & {\rotatebox{90}{\textbf{Photorealistic}}} &  & {\rotatebox{90}{\textbf{Qualitative}}} & {\rotatebox{90}{\textbf{User Study}}} & 
            {\rotatebox{90}{\textbf{\begin{tabular}[c]{@{}l@{}}Quantitative\\
            Metrics\end{tabular}}}}\\ 
    \midrule 
    
     2016  & Gatys~\etal \cite{gatys2015neural} & \checkmark & \checkmark & 
     & &  \\ \cline{1-4} \cline{6-8}

    2017 & Gatys~\etal \cite{gatys2017controlling} &  \checkmark & & 
    &  & \\ \cline{1-4} \cline{6-8}
   
    2017 & Li~\etal\cite{Li2017demistifying} & \checkmark &  & & 
    & & \\ \cline{1-4} \cline{6-8}
    
    2017 & Risser~\etal \cite{Risser2017stable} &  \checkmark & & &
    \checkmark & & Speed \\ \cline{1-4} \cline{6-8}
    
    2017 & Li~\etal \cite{Li2017laplacian} &  \checkmark & & & 
    \checkmark & & Content loss \& Style loss \\ \cline{1-4} \cline{6-8}
    
    2017 & Li and Wand \cite{li2016combining} & \checkmark & \checkmark &  &
   \checkmark & \centering I &  \\ \cline{1-4} \cline{6-8}
    
    2017 & Luan~\etal \cite{Luan2017deep} &  & \checkmark &  & 
    \checkmark & \checkmark &  \\ \cline{1-4} \cline{6-8}

    2017 & Mechrez~\etal \cite{mechrez2017photorealistic} &  & \checkmark &  & 
    \checkmark & \checkmark & Speed \\ \cline{1-4} \cline{6-8}
    
    2019 & \begin{tabular}[c]{@{}l@{}}Penh{\"o}uet and\\ Sanzenbacher \cite{penhouet2019automated}\end{tabular}  &  
    & \checkmark & &
    & & \\

    \bottomrule
    \end{tabular}
    }
\end{table}

\subsection{Images}

\subsubsection{Online Image Optimization Neural Methods (Table~\ref{tbl1})}

Gatys~\etal \cite{gatys2015neural,gatys2016image} noticed that the powerful capabilities of CNNs can be utilized for more than just object classification. Observing that an input image can be transformed into feature maps that increasingly care about the content rather than any detail about the color or texture as we go deeper into a CNN, they proposed an image-optimization technique that faithfully replicates artistic qualities of artworks on real photographs. Their technique falls into the Image-Optimization-Based Online Neural Methods and attempts to minimize an objective function that encompasses definitions of content and style loss. The content loss is defined by the squared euclidean distance between the feature representations $F^{l} (l=conv4\_2)$ of the content image ($I_{c}$) and the target image ($I_{t}$):

\begin{equation} \label{eq:gatys-contentloss}
    \mathcal{L}_{content} = \frac{1}{2} \sum_{l}  \Big( F^{l}(I_t) - F^{l}(I_c) \Big)^{2}
\end{equation}

To represent the style of the input style image, a feature space is designed to capture the texture and color information. This space looks at spatial correlations within a layer of a network. Features are extracted from multiple layers ($conv1\_1$, $conv2\_1$, $conv3\_1$, $conv4\_1$, $conv5\_1$) and the feature correlations are given by the Gram matrix which contains non-localized information about the image. The resulting style loss is defined by the squared Euclidean distance between the Gram-based style representations of the style image $I_s$ and target image $I_t$:
\begin{equation} \label{eq:gatys-styleloss}
    \mathcal{L}_{style} = \gamma \sum_{l} w_{l} \Big( \mathcal{G}(F^{l}(I_t)) - \mathcal{G}(F^{l}(I_s)) \Big)^{2}
\end{equation}
where $\gamma$ is a constant that accounts for the number of values in each layer and $w_{l}$ are weighting factors for the contribution of each layer.

To generate the target image that renders the input content image artistically stylized based on the style of the input style image, the system then tries to minimize the loss function: 
\begin{equation} \label{eq:gatys-totalloss}
    \mathcal{L}_{total} = \alpha \mathcal{L}_{content} + \beta \mathcal{L}_{style}
\end{equation}
where $\alpha$ and $\beta$ are used to control the weight of the content and style components in the stylized result, with $\beta$ often being much larger. These values stay fixed over the stylization process. An extension of this algorithm was the introduction of style factorization that allows stylizing different regions of the image with different styles \cite{gatys2017controlling}. Other image-optimization-based online methods include the systems proposed by Risser~\etal \cite{Risser2017stable} and Li~\etal \cite{Li2017laplacian}, whereas the work of Li~\etal \cite{li2016combining} attempts to model the style in a non-parametric way utilizing a patch-based Markov Random Field (MRF) prior in their style loss. 


As these algorithms were implemented in the early period of NST research, no particular emphasis was given to evaluation. After the work of Gatys~\etal \cite{gatys2015neural,gatys2016image}, researchers started to provide Qualitative Evaluations in the form of visual side-by-side comparisons \cite{Risser2017stable,Li2017laplacian,li2016combining}. Risser~\etal \cite{Risser2017stable} included speed comparisons, Li~\etal \cite{Li2017laplacian} discussed content loss and style loss as a way of measuring stylization performance, and Li and Wand \cite{li2016combining} provided some insights from an informal user study.

\subsubsection{Offline Model Optimization Neural Methods}

Whilst Online Image Optimization NST techniques produce visually appealing results, they are inefficient in terms of speed and computational cost. Model-optimization-based offline Neural Methods were introduced to mitigate the slow optimization process. These techniques attempt to optimize a generative model offline and produce a stylized image faster, with a single forward pass through the trained model. Formally, these models learn a style transfer function $\varPhi_\mathrm{w}$ which can be expressed by a feed-forward network with parameters $\mathrm{w}$, and is optimized over a set of content images $I_c$ for one or more style images $I_s$:

\begin{equation} \label{eq:model-optimization}
    \mathrm{w}^* = \argmin_\mathrm{w} \mathcal{L}_{total}(I_c, I_s, \varPhi_{\mathrm{w}^*}(I_c))
\end{equation}

As described by Jing~\etal \cite{jing2019neural}, depending on the number of styles that the feed-forward network can reproduce, the Model Optimization methods can be distinguished into \textit{Per-Style-Per-Model (PSPM)} methods, \textit{Multiple-Style-Per-Model (MSPM)} methods, and \textit{Arbitrary-Style-Per-Model (ASPM)}. In this section, we introduce another category, \textit{Per-Artist-Per-Model (PAPM)}, in order to accommodate the methods that attempt to capture a particular artistic style or genre instead of emulating one particular input reference style image.

\subsubsection*{Per-Style-Per-Model Methods (Table~\ref{tbl2})}

Johnson~\etal \cite{johnson2016perceptual} was the first to address the inefficiency of the method of Gatys~\etal \cite{gatys2016image}. Although the optimization problem remains the same, their work utilizes perceptual loss functions to train their network. Consequently, they generate stylized images three orders of magnitude faster. Further improvements to the inference speed of the trained image transformation network were introduced by Ulyanov~\etal \cite{ulyanov2016texture}. Later, they published an improvement to their algorithm \cite{ulyanov2017improved} that achieves visually better results. This was based on the idea of replacing the Batch Normalization layers in the generator with Instance Normalization \cite{ulyanov2016instance}. 
The work by Li and Wand \cite{li2016precomputed} addresses the speed issue by training a Markovian feed-forward network that also reproduces one reference style.

More recently, researchers have considered how to preserve the depth and structural information of the content images when performing stylization. Models have been developed that attempt to alleviate the undesired effects that occur when the input images include numerous objects at various depths. The system of Liu~\etal \cite{liu2017depth} is built on the work of Johnson~\etal \cite{johnson2016perceptual} but introduces a depth reconstruction loss during training. Similarly, Cheng~\etal \cite{cheng2019structure} rely on depth maps and image edges to generate results of higher quality and retained structure. Other work combines the suggested depth loss \cite{liu2017depth} with an improved depth prediction network and with Instance Normalization layers for aesthetic and depth-preserving stylized images \cite{ioannou2022depth}.

\begin{table}[htb]
\setlength\extrarowheight{1pt}

\centering
    \caption{Per-Style-Per-Model Offline Model Optimization Methods. All methods are suitable for artistic style transfer.}\label{tbl2}

    \resizebox{1\linewidth}{!}{%
    
    \begin{tabular}{ p{2em}  p{10em} | p{1em} | p{1em}| p{1em} | p{1em} | p{1em} | p{12em}}

    \toprule
        \multicolumn{1}{l}{\textbf{Year}} & \multicolumn{1}{l}{\textbf{Method}} & \multicolumn{3}{l}{\textbf{Dataset}} & \multicolumn{3}{l}{\textbf{Evaluation}}   \\ 
    \midrule 
            \multicolumn{1}{c}{} & {} & {\rotatebox{90}{\textbf{MS COCO}}} & {\rotatebox{90}{\textbf{ImageNet}}} &  & {\rotatebox{90}{\textbf{Qualitative}}} & {\rotatebox{90}{\textbf{User Study}}} & 
            {\rotatebox{90}{\textbf{\begin{tabular}[c]{@{}l@{}}Quantitative\\
            Metrics\end{tabular}}}}\\ 
    \midrule 
    
     2016  & Johnson~\etal \cite{johnson2016perceptual} & \checkmark &  & & 
     \checkmark & & \begin{tabular}[c]{@{}l@{}}Content loss \& Style loss,\\Speed\end{tabular} \\ \hline

     2016 & Li and Wand \cite{li2016precomputed} &  & \checkmark &  &
     \checkmark & & Speed, Memory \\ \hline
     
     2016 & Ulyanov~\etal \cite{ulyanov2016texture}  &  & \checkmark &  &
     \checkmark & & Speed, Memory \\ \hline
     
     2017 & Ulyanov~\etal \cite{ulyanov2017improved}   &  \checkmark & \checkmark &  &
     \checkmark &  &    \\ \hline

     2017 & Liu~\etal \cite{liu2017depth} & \checkmark & & &  
     \checkmark & & \begin{tabular}[c]{@{}l@{}}Dept/Edge/Saliency map\end{tabular} \\ \hline
     
     2019 & Cheng~\etal \cite{cheng2019structure} & \checkmark & & &
     \checkmark & \checkmark & \begin{tabular}[c]{@{}l@{}}Speed, Memory, SSIM,\\Depth map, Edge map\end{tabular}  \\  \hline
      
     2022 & \begin{tabular}[c]{@{}l@{}}Ioannou and\\ Maddock~\cite{ioannou2022depth}\end{tabular} & \checkmark & & & 
     \checkmark & \checkmark & \begin{tabular}[c]{@{}l@{}}Depth map, Saliency map,\\ SSIM, Hist, aHash, dHash\end{tabular}  \\  
            
    \bottomrule
    \end{tabular}
    }
\end{table}

\subsubsection*{Per-Artist-Per-Model Methods (Table~\ref{tbl3})} 

Since the seminal work of Gatys~\etal \cite{gatys2016image}, researchers have attempted to view and tackle the style transfer problem from different perspectives. Some approaches have tried to redefine what style is and how it can be captured and reproduced, while other methods address the aesthetics of the results. The system proposed by Sanakoyeu~\etal \cite{sanakoyeu2018style} avoids fixed style representations and focuses on the details that are relevant to the style when measuring the similarity in content between the input and the stylized image. This more generalized procedure is capable of transferring not only the style of one particular painting but also the distinctive style of \textit{Cezanne} or \textit{Van Gogh}. In a similar spirit, Kontovenko~\etal \cite{kotovenko2019content} propose a fixpoint triplet style loss (to learn nuanced variations within a style)  and a disentanglement loss (to prohibit the influence of the content on the stylization), and extract style from a group of examples of the same overall style (but with subtle variations) in order to generate stylizations in the style of \textit{Picasso} or \textit{Kandinsky}. 

\begin{table}[htb]
\setlength\extrarowheight{1pt}

\centering
    \caption{Per-Artist-Per-Model Offline Model Optimization Methods. The methods are suitable for transferring a particular style, or the style of a specific art genre/artist.}\label{tbl3}

    \resizebox{1\linewidth}{!}{%
    
    \begin{tabular}{ p{2em}  p{12em} | p{5em} | p{1em} | p{1em} | p{2em} | p{10em}}

    \toprule
        \multicolumn{1}{l}{\textbf{Year}} & \multicolumn{1}{l}{\textbf{Method}} & \multicolumn{1}{l}{\textbf{Dataset}} & \multicolumn{4}{l}{\textbf{Evaluation}}   \\ 
    \midrule 
            \multicolumn{1}{c}{} & \multicolumn{1}{c}{} &   & \centering{\rotatebox{90}{\textbf{Qualitative}}} & \centering{\rotatebox{90}{\textbf{User Study}}} & \centering{\rotatebox{90}{\textbf{\begin{tabular}[c]{@{}l@{}}User Study \\w/ Experts\end{tabular}}}} & {\rotatebox{90}{\textbf{\begin{tabular}[c]{@{}l@{}}Quantitative\\
            Metrics\end{tabular}}}} \\ 
    \midrule 

     2018 &  Sanakoyeu~\etal \cite{sanakoyeu2018style} & \multirow{3}{10em}{\begin{tabular}[c]{@{}l@{}}Places365 \&\\ Wikiart\end{tabular}} & 
     \centering \checkmark &  & \centering \checkmark & \begin{tabular}[c]{@{}l@{}}Deception Rate,\\ Speed, Memory\end{tabular}  \\ \cline{1-2} \cline{4-7}
     
    2019 & Kotovenko~\etal \cite{kotovenko2019content} &  & 
    \centering \checkmark & \centering \checkmark & \centering \checkmark & \begin{tabular}[c]{@{}l@{}}Deception Rate,\\ Style Divergence ($D_{KL}$)\end{tabular} \\  

        \bottomrule
    \end{tabular}
    }
\end{table}


\subsubsection*{Multiple-Style-Per-Model Methods (Table~\ref{tbl4})} 

An improvement to the style transfer algorithms was to create models capable of reproducing more than one style per trained network. Dumoulin~\etal's method \cite{dumoulin2017learned} reduces style images into points in an embedding space and employs conditional instance normalization, allowing the style transfer network to learn multiple styles. Chen~\etals~\cite{chen2017stylebank} \textit{StyleBank} decouples content and style representations utilizing multiple convolution filter banks with each filter bank representing one style, making the method capable of producing stylizations of more than one style or even new style fusion effects. Other notable work that achieves multi-style transfer includes Zhang and Dana \cite{zhang2018multi} and the technique of Li~\etal \cite{li2017diversified} on texture synthesis. 
\begin{table}[htb]
\setlength\extrarowheight{1pt}
\centering
    \caption{Multiple-Style-Per-Model Offline Model Optimization Methods. The methods are suitable for artistic style transfer.}\label{tbl4}

   \resizebox{\linewidth}{!}{%
   
   \begin{tabular}{ p{2em} p{12em} | p{5em}| p{2em}| p{1em} | p{1em} | p{1em} | p{8em}}

    \toprule
        \multicolumn{1}{l}{\textbf{Year}} & \multicolumn{1}{l}{\textbf{Method}} & \multicolumn{3}{l}{\textbf{Dataset}} & \multicolumn{3}{l}{\textbf{Evaluation}}   \\ 
    \midrule 
            \multicolumn{1}{c}{} & {} & \centering{\rotatebox{90}{\textbf{Content}}} & \centering{\rotatebox{90}{\textbf{No. Styles}}} &  & {\rotatebox{90}{\textbf{Qualitative}}} & {\rotatebox{90}{\textbf{User Study}}} & 
            {\rotatebox{90}{\textbf{\begin{tabular}[c]{@{}l@{}}Quantitative\\
            Metrics\end{tabular}}}} \\ 
    \midrule 

     2017 &  Dumoulin~\etal~\cite{dumoulin2017learned} & ImageNet  & 32  &  & & \\ \hline
     
     2017 & Chen~\etal~\cite{chen2017stylebank} & MS COCO & 50 & 
     & \checkmark & &  \\ \hline
    
     2017 & Li~\etal~\cite{li2017diversified} & DTD & 1000 &
     & \checkmark & & \\ \hline

     2018 & Zhang and Dana \cite{zhang2018multi} & MS COCO & 1000 & 
      & \checkmark & & Speed, Memory  \\   

        \bottomrule
    \end{tabular}
    }
\end{table}


\subsubsection*{Arbitrary-Style-Per-Model Methods (Table~\ref{tbl5} \& Table~\ref{tbl6})} 

These NST methods attempt to create one model capable of transferring arbitrary styles. For example, the algorithm of Chen and Schmidt \cite{chen2016fast} encompasses a procedure (\textit{``Style Swap"}) in which the content image is replaced patch-by-patch by the style image after a set of activation patches is extracted both from content and style activations. Later, the Adaptive Instance Normalization layer (AdaIN) \cite{huang2017arbitrary} was suggested:

\begin{equation}
    \begin{aligned}
        AdaIN(F(I_{c}), F(I_{s})) = \sigma(F(I_{s})) \Big(\frac{F(I_{c}) - \mu(F(I_{c}))}{\sigma(F(I_{c}))}\Big)\\ + \mu(F(I_{s}))
    \end{aligned}
\end{equation}
where $F$ stands for the feature activations. This allows transferring the channel-wise mean and variance feature statistics between the content and style feature activations, achieving arbitrary style transfer. Their algorithm is also significantly faster than previous approaches. Following this, a myriad of approaches have been published that propose arbitrary-style-per-model methods, improving upon performance \cite{ghiasi2017exploring,li2017universal,xu2018learning,huo2021manifold,park2019arbitrary,svoboda2020two,an2020ultrafast} or addressing a particular use case, e.g., aesthetics of the results \cite{hu2020aesthetic,hong2023aespa} or geometric warping \cite{liu2021learning}. Approaches have also emerged that utilize meta networks \cite{shen2018neural} or more recently transformers and attention \cite{park2019arbitrary,liu2021adaattn,deng2022stytr2,luo2022consistent,zhu2023all}. Other more recent algorithms propose the use of contrastive learning \cite{zhang2023unified}, quantization \cite{huang2023quantart} and image restoration \cite{ma2023rast}.

Arbitrary NST has also been exploited for the generation of photorealistic stylizations \cite{li2018closed,yoo2019photorealistic,an2020ultrafast,ke2023neural}. These works attempt to stylize an input content photo using a reference photo as a style image. The aim is to avoid distortions and visual artifacts when transferring the photographic style to the input content photo. Some of the methods included in Table~\ref{tbl5} (\cite{huo2021manifold,huang2023quantart}) are also capable of performing photorealistic style transfer.

\begin{table}[htb]

\setlength\extrarowheight{0.5pt}
\centering
 
    \caption{Arbitrary-Style-Per-Model Offline Model Optimization Methods for Artistic NST.}\label{tbl5}
  

    \resizebox{1\linewidth}{!}{%
    
    \begin{tabular}{ p{2em} | p{10.5em} | p{1em} | p{1em}| p{7em} | p{1em} | p{1em} | p{9em}}

    \toprule
        \multicolumn{1}{l}{\textbf{Year}} & \multicolumn{1}{l}{\textbf{Method}} & \multicolumn{3}{l}{\textbf{Dataset}} & \multicolumn{3}{l}{\textbf{Evaluation}}   \\ 
    \midrule 
            \multicolumn{1}{c}{} & {} & {\rotatebox{90}{\textbf{MS COCO}}} & {\rotatebox{90}{\textbf{Wikiart}}} & {\rotatebox{90}{\textbf{Other}}} & {\rotatebox{90}{\textbf{Qualitative}}} & {\rotatebox{90}{\textbf{User Study}}} & 
            {\rotatebox{90}{\textbf{\begin{tabular}[c]{@{}l@{}}Quantitative\\
            Metrics\end{tabular}}}}\\ 
    \midrule 
    
        2016 & \begin{tabular}[c]{@{}l@{}}Chen and\\ Schmidt~\cite{chen2016fast}\end{tabular} & 
        \checkmark & \checkmark &  &
        \checkmark &  & Speed  \\ \hline

        \multirow{4}{0pt}{2017} & 
        \begin{tabular}[c]{@{}l@{}}Huang and\\ Belongie~\cite{huang2017arbitrary}\end{tabular} & 
        \checkmark &  \checkmark &  & 
        \checkmark & & \begin{tabular}[c]{@{}l@{}}Content \& Style Loss,\\ Speed\end{tabular} \\ \cline{2-8}

         & Ghiasi~\etal \cite{ghiasi2017exploring} & 
        &   &  \begin{tabular}[c]{@{}l@{}}ImageNet \&\\ PBN, TDT\end{tabular} & 
        \checkmark &  &  \\ \cline{2-8}
        
         & Li~\etal \cite{li2017universal} & 
        \checkmark &   &  & 
        \checkmark & & Style Loss, Speed \\ \hline

        \multirow{5}{0pt}{2018} & Gu~\etal \cite{gu2018arbitrary} & 
        & & ImageNet & 
        \checkmark & \checkmark & Speed  \\ \cline{2-8}
           
         & Shen~\etal \cite{shen2018neural} &  
        \checkmark & \checkmark &  &
        \checkmark &  & Speed, Memory    \\ \cline{2-8}
        
         & Xu~\etal \cite{xu2018learning} & 
        & & \begin{tabular}[c]{@{}l@{}}Behance, DTD,\\ CelebA,\\ UCSD birds, cars, \\Oxford Building\end{tabular} 
        & \checkmark & \checkmark &  \\ \hline

        \multirow{2}{0pt}{2019} & Li~\etal \cite{li2019learning} & 
        \checkmark & \checkmark &  & 
        \checkmark & \checkmark & Speed  \\ \cline{2-8}
        
          & Park and Lee~\cite{park2019arbitrary} & 
        \checkmark & \checkmark &  & 
        \checkmark & \checkmark & Speed \\ \hline

        \multirow{2}{0pt}{2020} & Hu~\etal \cite{hu2020aesthetic} & 
        \checkmark & \checkmark &  & 
        \checkmark & \checkmark &  \\ \cline{2-8}
        
         & Svodoba~\etal \cite{svoboda2020two} & 
         & \checkmark & Places365 &
        \checkmark &  &  \\ \hline

        \multirow{9}{0pt}{2021} & Liu~\etal \cite{liu2021adaattn} & 
        \checkmark & \checkmark &  &  
        \checkmark & \checkmark &  Speed  \\ \cline{2-8}
        
         & An~\etal \cite{an2021artflow} & 
        \checkmark & \checkmark &  Metfaces &
        \checkmark & \checkmark &  \begin{tabular}[c]{@{}l@{}} SSIM,\\ Content \& Style Error,\\ Content Leak, Speed\end{tabular}  \\ \cline{2-8}

          & Liu~\etal \cite{liu2021learning} & 
        \checkmark &  &  PF-PASCAL & 
        \checkmark & \checkmark & Speed  \\ \cline{2-8}

         & Huo~\etal \cite{huo2021manifold} & 
        \checkmark & \checkmark &  & 
        \checkmark & \checkmark &  Speed \\ \cline{2-8}

         & Liu and Zhu \cite{liu2021structure} & 
        \checkmark & \checkmark &  &
        \checkmark & \checkmark &  \begin{tabular}[c]{@{}l@{}}Depth map, Edge map,\\ Saliency map, SSIM,\\Hist, aHash, dHash\end{tabular}    \\ \cline{2-8}

        & \textbf{Chen~\etal \cite{chen2021artistic}} & 
        \checkmark & \checkmark &  &
        \checkmark & \checkmark & Speed    \\ \hline

        \multirow{4}{0pt}{2022} & Deng~\etal \cite{deng2022stytr2} & 
        \checkmark & \checkmark &  &
        \checkmark & \checkmark  &\begin{tabular}[c]{@{}l@{}}SSIM,\\ Content \& Style Error,\\ Content Leak, Speed\end{tabular}   \\ \cline{2-8}
        
         & Luo~\etal \cite{luo2022consistent} & 
        \checkmark & \checkmark &  &
        \checkmark & \checkmark & Speed \\ \cline{2-8}

         & \textbf{Wang~\etal \cite{wang2022aesust}} & 
        \checkmark & \checkmark &  &
        \checkmark & \checkmark & CF, GE, LP, Speed \\ \hline
        
        2023 & Ruta~\etal \cite{ruta2023neat} & 
       \checkmark & \checkmark &  &
       \checkmark & \checkmark & \begin{tabular}[c]{@{}l@{}}Chamfer, LPIPS,\\ SFID, Speed\end{tabular}   \\ \hline

       2023 & Huang~\etal \cite{huang2023quantart} & 
       \checkmark & \checkmark & \begin{tabular}[c]{@{}l@{}}FFHQ \\MetFaces \\ LandscapesHQ \end{tabular} &
       \checkmark & \checkmark & \begin{tabular}[c]{@{}l@{}}LPIPS, Style Error\\ FID, ArtFID, Speed\end{tabular}   \\ \hline

        2023 & Xu~\etal \cite{xu2023learning} & 
       \checkmark & \checkmark &  &
       \checkmark & \checkmark & \begin{tabular}[c]{@{}l@{}}LPIPS, Style Error\\ Speed\end{tabular}   \\ \hline

        2023 & Tang~\etal \cite{tang2023master} & 
       \checkmark & \checkmark &  &
       \checkmark & \checkmark & \begin{tabular}[c]{@{}l@{}}Content \& Style Error,\\ $\mathcal{L}_{sim}$\end{tabular}   \\ \hline

        2023 & Hong~\etal \cite{hong2023aespa} & 
       \checkmark & \checkmark &  &
       \checkmark & \checkmark & \begin{tabular}[c]{@{}l@{}}Style Error,\\ CF, LP, Speed\end{tabular}   \\ \hline

       2023 & Zhu~\etal \cite{zhu2023all} & 
       \checkmark & \checkmark &  &
       \checkmark & \checkmark & \begin{tabular}[c]{@{}l@{}}Content \& Style Error,\\ LPIPS, Speed\end{tabular}   \\ \hline

        2023 & Gu~\etal \cite{gu2023two} & 
       \checkmark & \checkmark &  &
       \checkmark & \checkmark & \begin{tabular}[c]{@{}l@{}}Content \& Style Error,\\ color loss, Speed\end{tabular}   \\ \hline

       2023 & Ma~\etal \cite{ma2023rast} & 
       \checkmark & \checkmark &  &
       \checkmark & \checkmark & \begin{tabular}[c]{@{}l@{}}Content \& Style Error,\\ LPIPS, Speed\end{tabular}   \\ \hline

       2023 & Li~\etal \cite{li2023frequency} & 
       \checkmark & \checkmark &  &
       \checkmark & \checkmark & \begin{tabular}[c]{@{}l@{}}Content \& Style Error,\\ SSIM, Speed\end{tabular}   \\ \hline

       2023 & Zhang~\etal \cite{zhang2023unified} & 
        & \checkmark & Places365  &
       \checkmark & \checkmark & \begin{tabular}[c]{@{}l@{}}Content Error, LPIPS\\ Deception rate\end{tabular}   \\

    \bottomrule
    \end{tabular}%
    }
\end{table}

\begin{table}[htb]
\setlength\extrarowheight{1pt}

 \centering
    \caption{Arbitrary-Style-Per-Model Offline Model Optimization Methods for Photorealistic Style Transfer.}\label{tbl6}

     \resizebox{\linewidth}{!}{%
     
     \begin{tabular}{ p{2em}  p{9em} | p{6em} | p{1em} | p{1em} | p{12em}}

    \toprule
        \multicolumn{1}{l}{\textbf{Year}} & \multicolumn{1}{l}{\textbf{Method}} & \multicolumn{1}{l}{\textbf{Dataset}} & \multicolumn{3}{l}{\textbf{Evaluation}}   \\ 
    \midrule 
            \multicolumn{1}{c}{} & \multicolumn{1}{c}{} &   & {\rotatebox{90}{\textbf{Qualitative}}} & {\rotatebox{90}{\textbf{User Study}}} & 
            {\rotatebox{90}{\textbf{\begin{tabular}[c]{@{}l@{}}Quantitative\\
            Metrics\end{tabular}}}}\\ 
    \midrule 

       
       2018 & Li~\etal\cite{li2018closed} & \multirow{4}{5.5em}{\centering MS COCO}& 
       \checkmark & \checkmark & Speed \\ \cline{1-2} \cline{4-6}
       
       2019 & Yu~\etal\cite{yoo2019photorealistic}  &  &
       \checkmark & \checkmark & \begin{tabular}[c]{@{}l@{}}Style loss, SSIM,\\ Speed, Memory\end{tabular} \\ \cline{1-2} \cline{4-6}
    
        2020 & An~\etal\cite{an2020ultrafast} &   &
        \checkmark & \checkmark & \begin{tabular}[c]{@{}l@{}}SSIM (edges, whole), Speed\end{tabular}   \\ \hline

        2023 & Ke~\etal\cite{ke2023neural} &   &
        \checkmark & \checkmark & \begin{tabular}[c]{@{}l@{}}SSIM, Style similarity \\ Speed\end{tabular}   \\
    \bottomrule
    \end{tabular}%
    }
\end{table}


\subsection{Videos}

The main challenge in stylizing a video sequence instead of just a single still image is achieving temporal consistency. Temporal incoherence is observed visually as flickering
between consecutive frames. This results in inconsistent stylization of moving objects (or idle objects but moving camera, or both moving) across subsequent frames. Ruder~\etal \cite{ruder2016artistic} proposed the first video style transfer algorithm that is based on the stylization algorithm by Gatys~\etal \cite{gatys2016image}. The algorithm utilizes a temporal constraint which penalises deviations along point trajectories in order to preserve a smooth transition between adjacent frames. Optical flow is considered and smoothed stylized videos are produced. The technique is based on the idea of initialising the style optimization algorithm for frame $i+1$ with the previous stylized frame warped:
\begin{equation} \label{eq:temporal-constraint}
    {x'}^{(i+1)} = \omega_{i}^{(i+1)} (x^{(i)})
\end{equation}
where $x^{(i)}$ denotes the stylized frames to be generated, and $\omega_{i}^{(i+1)}$ is the function that takes as input an image and warps it using the optical flow field that was calculated between frames $i$ and $i+1$ of the original video. A stronger consistency is achieved by additionally detecting disoccluded regions and motion boundaries. The approach was later improved and extended to spherical images and videos \cite{ruder2018artistic}, hence being applicable for Virtual Reality (VR) applications.

Subsequent studies attempt to improve the speed of computation \cite{huang2017real,gao2018reconet}, whilst others account for the stylization quality \cite{sanakoyeu2018style} or the depth-preserving capabilities of the trained network \cite{liu2021structure,ioannou2023depth}. In addition, Gao~\etal \cite{gao2020fast} proposed a fast model that incorporates multiple styles. The more recent approaches, though, are capable of arbitrary style transfer to video sequences. Some of them are extensions to image stylization with additional temporal considerations \cite{li2019learning,liu2021adaattn,liu2021structure,gu2023two} whilst others are explicitly focused on video stylization \cite{wang2020consistent,deng2021arbitrary,lu2022universal,wu2022ccpl}. A photorealistic approach to video stylization has also been suggested \cite{xia2021real}. Table~\ref{tbl7} summarises the video stylization approaches and the evaluation techniques each utilized. Although not included in Table~\ref{tbl7}, it is worth mentioning that other studies have also emerged that try to stylize different object classes with a different style \cite{kurzman2019classbased} or support user controllability \cite{Jamriska19-SIG}.

\begin{table}[htb]

\centering
    \setlength\extrarowheight{1.5pt}
    \caption{Video NST Methods. The table includes image NST methods from Table~\ref{tbl5} that have also been trained to work for videos. Some of these approaches have conducted a User Study but there are cases in which the study does not assess the performance of stylized videos e.g., \cite{liu2021adaattn,sanakoyeu2018style,chen2021artistic}.}
    \label{tbl7}
       
    \resizebox{1\linewidth}{!}{%
    
    \begin{tabular}{ p{2em}  p{10em} | p{1em} | p{1em}| p{7em} | p{1em} | p{1em} | p{10.5em}}

    \toprule
        \multicolumn{1}{l}{\textbf{Year}} & \multicolumn{1}{l}{\textbf{Method}} & \multicolumn{3}{l}{\textbf{Dataset}} & \multicolumn{3}{l}{\textbf{Evaluation}}   \\ 
    \midrule 
            \multicolumn{1}{c}{} & {} & {\rotatebox{90}{\textbf{MS COCO}}} & {\rotatebox{90}{\textbf{Wikiart}}} & {\rotatebox{90}{\textbf{Other}}} & {\rotatebox{90}{\textbf{Qualitative}}} & \centering {\rotatebox{90}{\textbf{User Study}}} & 
            {\rotatebox{90}{\textbf{\begin{tabular}[c]{@{}l@{}}Quantitative\\
            Metrics\end{tabular}}}}\\ 
    \midrule 

     \multicolumn{8}{l}{\small{\textit{Online Frame-By-Frame Optimization}}} \\  \hline 

    2016 & Ruder~\etal\cite{ruder2016artistic} &  & & & \checkmark & & Speed, Warping error \\ \hline
     
     \multicolumn{8}{l}{\small{\textit{Per-Style-Per-Model (Offline Model Optimization)}}} \\  \hline 
    
      2017 & Huang~\etal\cite{huang2017real} & & & Videvo.net & 
      \checkmark & \centering \checkmark & Speed, Warping error \\ \hline
    
      2018 & Ruder~\etal\cite{ruder2018artistic} & \checkmark & & Hollywood2 &
      \checkmark & \centering \checkmark & Speed, Warping error \\ \hline

     2018 & Sanakoyeu~\etal\cite{sanakoyeu2018style} & & \checkmark & Places365  & 
     \checkmark & \centering  & Deception Rate, Speed, Memory \\ \hline
     
     2018 & Gao~\etal \cite{gao2018reconet}  & & & \begin{tabular}[c]{@{}l@{}}FlyingThings3D,\\ Monkaa\end{tabular} & 
     \checkmark & \centering \checkmark & Speed, Warping error \\ \hline
     
     2023 & \begin{tabular}[c]{@{}l@{}}Ioannou and\\ Maddock~\cite{ioannou2023depth}\end{tabular} 
       & \checkmark & & \begin{tabular}[c]{@{}l@{}} FlyingThings3D,\\Monkaa \end{tabular} & 
       \checkmark & & \begin{tabular}[c]{@{}l@{}}Warping error,\\Depth loss, LPIPS\end{tabular} \\ 
   
    \bottomrule

     \multicolumn{8}{l}{\small{\textit{Multiple-Style-Per-Model (Offline Model Optimization)}}} \\ \hline
     
     2020 & Gao~\etal \cite{gao2020fast} & 
     \checkmark & & Videvo.net &
      \checkmark & \centering \checkmark & Speed, Warping error \\ 
     
     \bottomrule

      \multicolumn{8}{l}{\small{\textit{Arbitrary-Style-Per-Model (Offline Model Optimization)}}} \\ \hline 
       \multicolumn{8}{l}{\scriptsize{\textit{Artistic Style Transfer}}} \\ \hline 
      
      2019 & Li~\etal \cite{li2019learning} & \checkmark & \checkmark & &
      \checkmark & \centering \checkmark & Speed, Heatmaps\\ \hline

       2020 & Wang~\etal\cite{wang2020consistent} & \checkmark & \checkmark & PBN & 
       \checkmark & \centering \checkmark & Speed, Warping error \\ 
        \hline

     2021 & Liu~\etal\cite{liu2021adaattn} & \checkmark & \checkmark & & 
     \checkmark & \centering & Speed, Warping error\\ 
     \hline

    2021 & Deng~\etal \cite{deng2021arbitrary} & \checkmark & \checkmark & & 
    \checkmark & \centering \checkmark & \begin{tabular}[c]{@{}l@{}}Content \& Style loss,\\ Temporal Mean \& Variance\end{tabular} \\  \hline

    2021 & Liu and Zhu~\cite{liu2021structure} & \checkmark & \checkmark &  & 
    \checkmark &  & \begin{tabular}[c]{@{}l@{}}Depth/Edge/Saliency map,\\SSIM, Hist, aHash, dHash\end{tabular} \\ \hline

      2021 & \textbf{Chen~\etal \cite{chen2021artistic}} & \checkmark & \checkmark &  & 
    \checkmark &  & LPIPS \\ \hline

     2022 & Lu and Wang~\cite{lu2022universal} & \checkmark & \checkmark & & 
     \checkmark &  & \begin{tabular}[c]{@{}l@{}}Content \& Style error, $D^{*}$,\\Speed, Heatmaps, LPIPS\textsubscript{vid} \end{tabular} \\  \hline

      2022 & Wu~\etal\cite{wu2022ccpl} & \checkmark & \checkmark & & 
      \checkmark & \centering \checkmark & \begin{tabular}[c]{@{}l@{}}Speed, Warping error\\SFID, LPIPS\textsubscript{vid}\end{tabular} \\  \hline 

      2023 & Gu~\etal\cite{gu2023two} &  & \checkmark & MPI Sintel & 
      \checkmark & \centering \checkmark & \begin{tabular}[c]{@{}l@{}}Speed, Warping error\\ LPIPS\textsubscript{vid}\end{tabular} \\ \hline 

      2023 & Zhang~\etal\cite{zhang2023unified} &  & \checkmark & Places265 & 
      \checkmark & & Warping error \\

      \bottomrule

     \multicolumn{8}{l}{\scriptsize{\textit{Photorealistic Style Transfer}}} \\ \hline
     
      2017 & Xia~\etal\cite{xia2021real} &  & & DAVIS 2017 & 
      \checkmark & \centering \checkmark & \begin{tabular}[c]{@{}l@{}}Speed, Warping error, TCC\end{tabular} \\   
    
    \bottomrule    
    
    \end{tabular}
    }
\end{table}

In addition to the evaluation techniques used in image style transfer, maintaining temporal consistency and avoiding incongruities and undesired effects is also important when evaluating video NST algorithms.
This is mainly measured with the help of optical flow (Warping error), or by looking at the perceptual (LPIPS) or depth (TCC) differences between consecutive frames. Qualitative evaluations and user studies are also used for the evaluation of the temporal aspect of the stylized results.

\subsection{Summary}

The NST literature encompasses a diverse range of techniques applied to both images and videos. However, despite the success of the field, a standardized benchmark evaluation procedure is yet to be established. Tables~\ref{tbl1}-\ref{tbl7} show some similarities in the evaluation approaches that have been used, but they also highlight the dissimilarities and divergences in the evaluation procedures employed across NST studies.

\begin{figure}[htb]
    \centering
    \includegraphics[width=\linewidth]{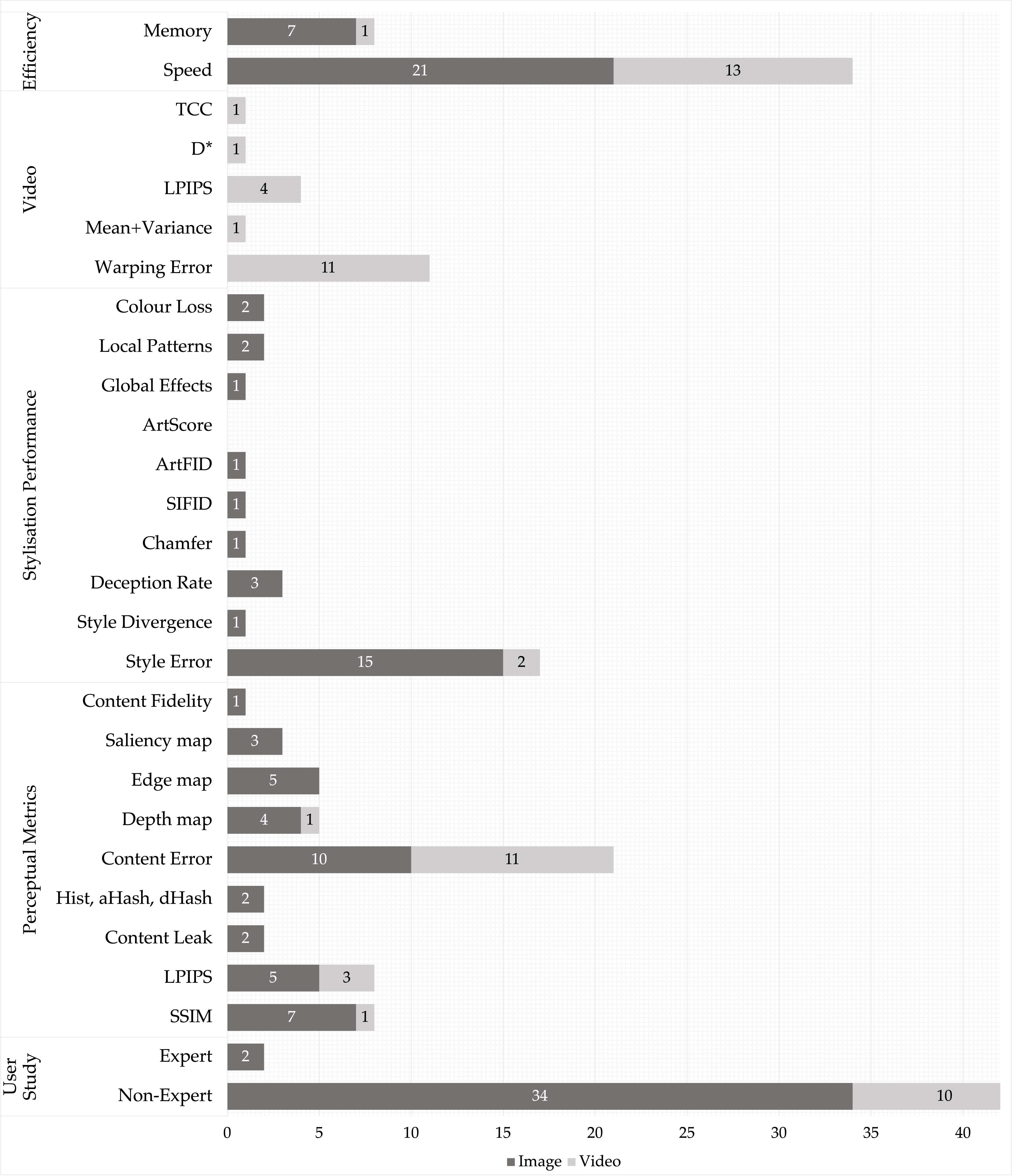}
    \caption[Quantitative Metrics Utilization]{The utilization of Quantitative Computational Metrics and User Studies by the NST methods. The graph shows the number of Image and Video NST methods that employed each of the quantitative metrics to evaluate their results and the amount of methods that conducted User Studies with Expert and Non-Expert participants. The graph shows that each method employs only a small selection of quantitative metrics for evaluative comparisons; some metrics are utilized more than others. At the time of this survey, the recently proposed \textit{ArtScore} \cite{chen2023learning} metric has not been utilized by any of the reviewed approaches.}
     \label{fig:quantitative_metrics_utilization}
\end{figure}

Figure~\ref{fig:quantitative_metrics_utilization} shows the frequency at which each metric is utilized by NST studies. Clearly, non-expert user studies dominate. 
The next section focuses on evaluation techniques.


\section{Current Evaluation Techniques}
\label{sec:current-evaluation}

\begin{figure}[htb]
\centering
\resizebox{1\linewidth}{!}{%
\begin{tikzpicture}[
    level 1/.style = {black, sibling distance = 1.7cm, level distance = 2cm},
  level 2/.style = {black, level distance = 1.9cm, sibling distance = 1.5cm, font=\bfseries},
  edge from parent fork down,
  level 3/.style = {black, level distance = 2cm, sibling distance=1.2cm, anchor=west,grow via three points={one child at (0.5,-0.65) and two children at (0.5,-0.7) and (0.5,-1.7)}, font=\bfseries,
    edge from parent path={(\tikzparentnode\tikzparentanchor) |- (\tikzchildnode\tikzchildanchor)}},
    every node/.append style = {draw}]
     
\node {\Large Evaluation Techniques}
  child {node[style=blue] {\begin{tabular}[c]{@{}l@{}}\large Qualitative\\ \large Evaluation\end{tabular}}
    child { 
        child {node[draw=none] {\begin{tabular}[c]{@{}l@{}}Visual\\ Side-by-Side\\ Comparisons\end{tabular}}}
    }
  }
  child [missing] {}
  child {node[style=blue] {\begin{tabular}[c]{@{}l@{}}\large Human\\ \large Evaluation\\ \large Studies\end{tabular}}
    child { 
    child {node[draw=none] {Non-Expert}} 
    child {node[draw=none] {Expert}} 
    }
  }
  child [missing] {}
  child [missing] {}
  child [missing] {}
  child [missing] {}
  child {node[style=blue] {\begin{tabular}[c]{@{}l@{}}\large Quantitative\\\large Metrics\end{tabular}}
    child {node[] {\begin{tabular}[c]{@{}l@{}}Perceptual\\ Metrics\end{tabular}}
      child  {node[draw=none] {SSIM}}
      child {node[draw=none] {LPIPS}}
        child {node[draw=none] {\begin{tabular}[c]{@{}l@{}}Content\\ Leak\end{tabular}}}
        child {node[draw=none] {Hist}}
        child {node[draw=none] {\begin{tabular}[c]{@{}l@{}}Content\\ Error\end{tabular}}}
        child {node[draw=none] {Depth Map}}
        child {node[draw=none] {Edge Map}}
        child {node[draw=none] {Saliency Map}}
        child {node[draw=none] {\begin{tabular}[c]{@{}l@{}}Content\\ Fidelity\end{tabular}}}
    }
    child [missing] {}
    child {node[] {\begin{tabular}[c]{@{}l@{}}Stylization\\ Performance\end{tabular}}
      child {node[draw=none] {Style Error}}
      child {node[draw=none] {\begin{tabular}[c]{@{}l@{}}Style\\ Divergence\end{tabular}}}
      child {node[draw=none] {\begin{tabular}[c]{@{}l@{}}Deception\\ Rate\end{tabular}}}
      child {node[draw=none] {\begin{tabular}[c]{@{}l@{}}Global\\ Effects\end{tabular}}}
      child {node[draw=none] {\begin{tabular}[c]{@{}l@{}}Local\\ Patterns\end{tabular}}}
      child {node[draw=none] {\begin{tabular}[c]{@{}l@{}}Color\\ Loss\end{tabular}}}
      child {node[draw=none] {Art-FID}}
      child {node[draw=none] {SIFID}}
      child {node[draw=none] {Chamfer}}
      child {node[draw=none] {ArtScore}}
    }
    child [missing] {}
    child {node[] {Video}
      child {node[draw=none] {\begin{tabular}[c]{@{}l@{}}Warping\\Error\end{tabular}}}
      child {node[draw=none] {\begin{tabular}[c]{@{}l@{}}Mean\\ \& Variance\end{tabular}}}
      child {node[draw=none] {LPIPS\textsubscript{vid}}}
      child {node[draw=none] {D\textsuperscript{*}}}
      child {node[draw=none] {TCC}}
    }
    child [missing] {}
    child {node[style=] {Efficiency}
        child {node[draw=none] {Speed}}
        child {node[draw=none] {Memory}}
    }
  };
\end{tikzpicture}%
}
\caption{Evaluation Techniques utilized in the NST literature categorized into Qualitative Evaluation, Human Evaluation Studies, and Quantitative Metrics.
The blue highlighting reflects the standard categorization in the literature (Figure~\ref{fig:evaluation-general-categories}). 
}
\label{fig:evaluation-taxonomy}
\end{figure}
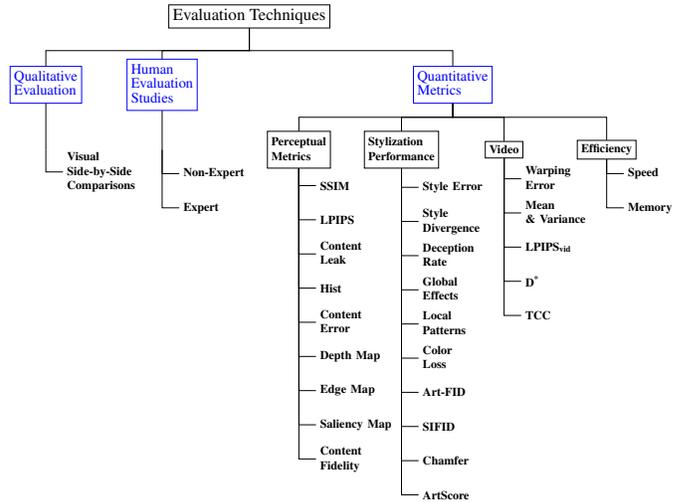

Informed by Tables~\ref{tbl1}-\ref{tbl7}, this section examines the evaluation methodologies utilized by the NST methods. 
Figure~\ref{fig:evaluation-taxonomy} depicts the categorization used. \emph{Qualitative Evaluation} involves subjective judgments based on side-by-side visual
comparisons provided by the author(s) of the relevant paper, \emph{Human Evaluation Studies} collect data from Expert and/or Non-Expert participants, and \emph{Quantitative Metrics} can be divided into Perceptual Metrics, Stylization Performance, Video Metrics, and Efficiency Metrics, depending on the aspect an automated metric aims to assess. Evaluation methodologies and respective datasets will be examined to shed light on the inconsistencies and limitations in evaluation. Ablation studies are also included. The final section presents and compares the results of state-of-the-art NST approaches.


\subsection{Datasets}
\label{sec:3-datasets}

As for all deep learning approaches, data used both in training and evaluation of NST methods exerts a notable influence on the results. 
Although for some areas of NST research (Photorealistic Image NST, Videos) there is a level of agreement regarding the data utilized for evaluation, in general, there exists a lack of consensus regarding the precise dataset to be used for all the different forms of evaluation that are conducted. Here, we attempt to investigate what data is utilized to evaluate and present the results of the current techniques, distinguishing between Image and Video NST. 

\subsubsection{Image NST Evaluation Data} 

Gatys~\etal \cite{gatys2016image} mainly demonstrated their technique by performing the stylization procedure on the photograph of Tuebingen Neckarfront by Andreas Praefcke (2003). In their showcase of synthesized results, multiple well-known artworks were employed as style images, e.g., The Shipwreck of the Minotaur by J.M.W. Turner (1805), and The Starry Night by Vincent van Gogh (1889). 
A similar selection of style images has been made by almost all NST researchers since then but with no particular agreement regarding the content images that are presented in the results.

Johnson~\etal \cite{johnson2016perceptual} stated that images from the MS COCO~\cite{lin2014microsoft} validation set were used for the evaluation. For quantitative evaluation, they used 50 content images. Other Per-Style-Per-Model (Table~\ref{tbl3}) and Multiple-Style-Per-Model (Table~\ref{tbl4}) approaches have mostly considered images that were present in previous methods \cite{ulyanov2016texture,li2016precomputed,chen2017stylebank,zhang2018multi}. Similarly, most Arbitrary-Style-Per-Model papers (Table~\ref{tbl5}) retrieve images from MS COCO for content and artworks from Wikiart \cite{mohammad2018wikiart} for style in their evaluation. Exceptions include the work of Gu~\etal \cite{gu2018arbitrary} which, in addition to content-style pairs from previous works, also used images from Ostragram \cite{ostagram}, the work of Li~\etal \cite{li2017universal} which utilized 40 images from \cite{karayev2014recognising}, and the work of Xu~\etal \cite{xu2018learning} which utilized the Behance test set \cite{wilber2017bam}. 

It is noteworthy that for Photorealistic Image NST (Table~\ref{tbl1}), the method of Luan~\etal \cite{Luan2017deep}, which was the first one chronologically, published an evaluation dataset composed of 60 content images and 60 style images. 
Subsequent studies for Photorealistic NST utilized this dataset in their evaluation \cite{mechrez2017photorealistic} and particularly for the user studies \cite{li2018closed,yoo2019photorealistic,an2020ultrafast}.

Nevertheless, for Artistic NST (Tables~\ref{tbl1}-\ref{tbl5}), the images present in both the Qualitative and Quantitative evaluations seem to be chosen at random with no particular agreement between the methods. Another point of disagreement is the number of stylized results (and the amount of content and styles that are used to produce them) that are utilized for all three evaluation types: \emph{Qualitative}, \emph{User Studies}, and \emph{Quantitative}. The same applies to the characteristics/attributes of the content and style images used to derive the results demonstrated in the evaluation, i.e., portrait photographs, landscapes, and lighting conditions. 

An effort to establish a benchmark dataset was made by Mould and Rosin \cite{mould2017developing}. They proposed a set of 20 images (used as content) that represent a range of possible subject matter and image features. Their approach discusses a range of principles and image characteristics for developing a suitable benchmark. More recently, Ruta~\etal \cite{ruta2023neat} proposed a large-scale dataset, namely \textit{BBST-4M}, composed of content and style images retrieved from Flickr and Behance.net, respectively. Utilizing a model capable of predicting if an image is stylistic or not, they filter out images from the two subsets, resulting in a final dataset of 2.2 million stylistic images and 2 million content images.

\subsubsection{Video NST Evaluation Data} 
\label{sec:video-nst-data}
The MPI Sintel \cite{Butler2012sintel} dataset is the most common source of data (for content videos) used in the evaluations of Video NST approaches (Table~\ref{tbl7}). As the dataset offers several video sequences and ground truth optical flow information, 
it is suitable for gauging the performance of the models in terms of temporal coherence and smoothness. For style images, there is no common practice, yet there is a presence of well-known artworks in the results. In some cases, frames from real-world videos (Videvo.net \cite{videvo2019}) or computer games (GTA \cite{richter2016playing}) are also used \cite{huang2017real,li2019learning,gao2020fast,ioannou2023depth}. 

The work of Xia~\etal \cite{xia2021real} on Photorealistic Video NST (Table~\ref{tbl7}) utilized the DAVIS 2017 \cite{pont20172017} validation dataset for evaluation. In contrast to the abundance of methods available for Photorealistic NST on still images, there is a scarcity of adequate approaches within the domain of Photorealistic Video NST. Consequently, it becomes challenging to agree on a standard dataset for their evaluation purposes.

\subsection{Qualitative Evaluation}

Qualitative evaluation in NST involves the visual inspection of the synthesized results. Researchers attempt to examine the generated images and derive judgements on their aesthetic quality and their ability to transfer the style efficiently while maintaining the contents.

Typically, qualitative evaluation is performed by demonstrating the results of the suggested method side-by-side with the results of similar or state-of-the-art approaches. The stylized images are also presented with the content and style images that were used to synthesize the results. Arguably the visual side-by-side comparison provided by all the NST approaches is not free from subjectivity. The authors attempt to elevate their results by comparing them to the results of state-of-the-art methods, commenting on visible artifacts of other methods and highlighting the effectiveness of their technique.

Image quality, perceptual quality and aesthetic appeal are some of the aspects that a standard qualitative evaluation addresses. Although the exact wording differs among the studies, a common practice has been the characterization of the results with comparative terms such as ``better", ``comparable", ``competitive" and ``slightly behind" when referring to the quality of the stylizations \cite{johnson2016perceptual,ulyanov2017improved,huang2017arbitrary}. Often, zoomed-in cut-outs from the stylized outputs are accompanied by comments that highlight the superiority of the suggested method in preserving fine-grained details or in applying stylization more effectively in detailed structures compared to the failure to do so of the methods in comparison \cite{sanakoyeu2018style,hu2020aesthetic}.

In other cases, the particular problem an approach attempts to solve is pointed out through these visual side-by-side comparisons. For example, structure-preserving \cite{cheng2019structure,liu2021structure} and depth-aware methods \cite{liu2017depth,ioannou2022depth} use the illustrations to emphasize the capabilities of the technique in retaining the 3D spatial layout and local structures of the content images, as well as the distance relationships between the different objects in the scene and the background-foreground contrast.

For Video NST, consecutive stylized frames are visually inspected. Qualitative evaluation is employed to visually identify incongruities and undesired artifacts and highlight how the proposed approach facilitates temporal stability. 


As stressed before, a significant point of disagreement lies in the selection of the style and content images that are used to generate the evaluation results. Every study showcases a diverse array of comparisons, and, due to the absence of a universally agreed-upon set of content and style images, it is possible that each method selectively employs specific content and style images that could potentially grant an advantage to their respective system. 
Examples of the most common content and style images used in qualitative evaluations are included in the supplementary material. Despite their extensive utilization, a standardized and universally accepted dataset has not been established. 
We argue that NST research and the community would benefit if studies used the same content and style images to derive their results for their qualitative comparisons. Such a practice could result in fairer comparisons and important insights for a better understanding of each method and its efficiency and quality.

\subsection{Human Evaluation Studies}
\label{sec:user-studies}

Due to the subjectivity in assessing stylizations, NST researchers commonly use Human Evaluation Studies to assess the performance of artistic/photorealistic image/video generation methods.  
As depicted in Tables~\ref{tbl1}-\ref{tbl7}, the majority of Image and Video NST approaches opt to conduct user studies to verify the effectiveness of the proposed system and further compare it with baselines. Despite the similarities in the way user studies are conducted, disparities arise in their design and formulation, and subsequently in the presentation of the results. A review of the NST papers reveals an array of factors that methods approach in varied ways, as summarized in Table~\ref{tab:user_studies_factors}.

\begin{table}[htb]
\centering
    \caption{The different factors in human evaluation studies}
    \label{tab:user_studies_factors}

    \begin{tabular}{p{0.15\linewidth} | p{0.75\linewidth}}
        \toprule
             \textbf{Factor} & \textbf{Description} \\ 
       \midrule
       
       Number of participants &  The number of participants, the number of responses collected per participant and/or the total votes. \\ \hline
        
        Participants’ profiles &  The age and sex of the participants, and whether the users are experts.  \\ \hline
        
        Data & The number of content and style images/videos used to derive the results shown to the participants and if samples are drawn from a larger evaluation set of content-style pairs (and how many). \\ \hline

        Presentation &   Whether the content and the style images are revealed in each question and whether all the stylizations of the methods in comparison are shown in each question or only two of them (the proposed method and one other selected randomly). \\ \hline
        
      Number of methods & The number of other methods that the proposed approach is compared against. \\ \hline

       Number of questions & The overall number of questions and the number of questions asked to each participant. \\ \hline

       Question formulation & What the participants are actually asked to answer and the exact factors they are asked to base their selection on. \\ \hline     
                
    \end{tabular}
    
\end{table}

    
    
    
    
    
     
    

\begin{figure}[htb]
    \centering
    \includegraphics[width=\linewidth]{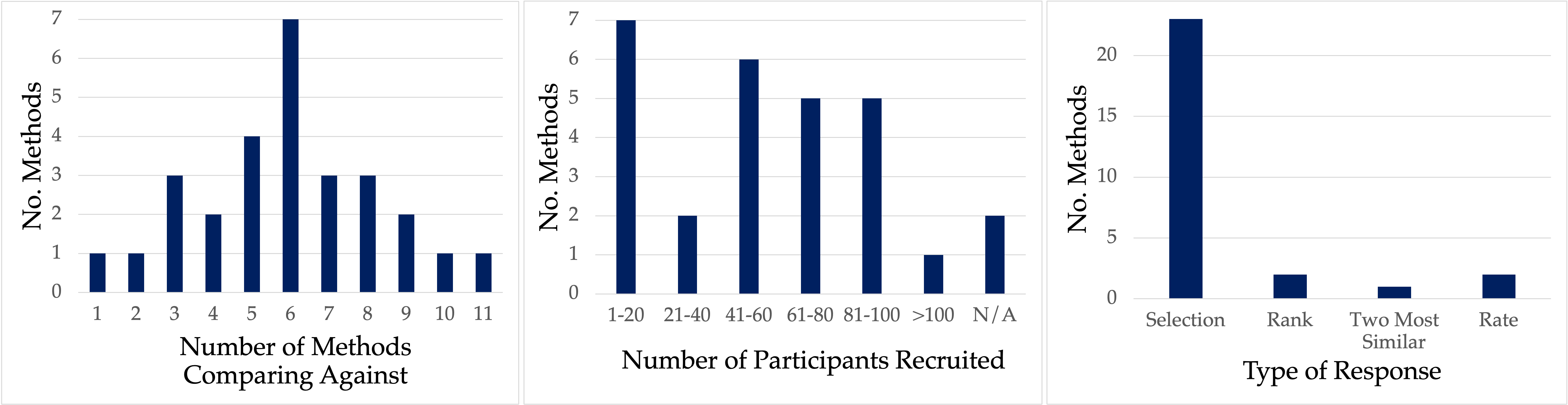}
    \caption[User Studies in NST]{The variation in methods comparing against, number of participants recruited, and response format in User Studies of Artistic Image NST approaches. ``N/A" in the second graph denotes that a study has not provided the relevant information.}
     \label{fig:user-studies-artistic-nst}
\end{figure}

To give an estimation of the variation in the settings of the user studies conducted in NST, Figure~\ref{fig:user-studies-artistic-nst} shows three graphs depicting the state of the literature. The graphs demonstrate how many methods (y-axis) choose to compare against different numbers of other state-of-the-art methods (Graph 1), the varied number of participants recruited by each NST study (Graph 2), and the different response formats that are followed (Graph 3). Although these graphs cannot be regarded as indicative of the whole NST literature (they only represent the Artistic Image NST methods reviewed in this paper), they provide a good approximation of the differences in the practices followed when designing and implementing user studies to assess stylized imagery.

\subsubsection{Number of Participants}
The number of participants recruited in the reviewed papers (Tables~\ref{tbl1}-\ref{tbl6}) varies from 10 to 220 people, even though on some occasions the number of subjects involved is omitted and only the number of votes is reported \cite{an2021artflow,an2020ultrafast}. In addition, it is essential to distinguish between expert (e.g., Art Historians \cite{sanakoyeu2018style,kotovenko2019content}) and non-expert participants. Sporadically, the age and sex of participants are provided \cite{cheng2019structure,deng2022stytr2,gu2023two}.

\subsubsection{Data generation/utilization}
Different practices are followed for the images/videos that are utilized in the user studies. Typically,
results are generated for an amount of content-style pairs, and for each method in comparison. 
In some cases, the number of content-style pairs chosen is not too large, and all the images are revealed to each participant. In other cases, a sample is drawn at random from the synthesized images, for example, 20 content-style pairs sampled from the total of 225 generated results (from 15 content and 15 style images) \cite{liu2021adaattn}. Another practice 
is to divide the repository of content-style pairs with the generated results into groups (e.g., a total of 240 outputs -- from 12 content and 20 style images -- divided into 5 groups, with each group containing 100 pairs \cite{hu2020aesthetic}) of which one is shown to each participant.

\subsubsection{Number of methods in comparison}
Depending on the particular use case and applicability of the proposed technique, or the specific problem it aims to solve, the number of methods that it is compared against is also varied. For Image NST, it can be from 1 to 11 approaches that are considered as baselines. The number of questions asked to each participant and the overall questions asked is another important parameter.

\subsubsection{Task presentation \& Question Formulation}
An essential part of the user study is the task presentation, and how each question is shown to the participant. This element in the design of the user study is varied amongst the NST research mainly by the following two factors: a) Number of synthesized images presented side-by-side, and b) Content \& style images display.
Multiple authors choose to compare the results of their approach with only one baseline each time \cite{wang2022aesust,tang2023master,hong2023aespa}, a design also referred to as \textit{two-alternative forced choice (2AFC), or A/B testing} \cite{kohavi2009controlled}. In most cases, though, participants are presented with three or more options simultaneously -- one output from each of the in-comparison methods. A few authors decided to reveal each generation result individually, asking for a rating \cite{Luan2017deep,huo2021manifold}. 
Another important consideration in NST user surveys is the decision to reveal to the user the content and style images that were used to derive the results. Even for a small group sample (20), it has been demonstrated that omitting the content and style images significantly influences the responses of the participants \cite{ioannou2022depth}.

The presentation of the questions is closely related to the formulation of the questions. The way the question is phrased and the type of response expected are impactful to the results. In most of the studies, the participants are asked to select their ``favorite" stylization (Figure~\ref{fig:user-studies-artistic-nst}, Graph 3). Examples of wordings used include ``most preferable", ``favorite", and ``the best" \cite{li2017universal,xu2018learning,park2019arbitrary,kitov2019depth,an2021artflow}. Occasionally, the participants are instructed to take into account indicators such as content presentation, style quality, and overall quality \cite{luo2022consistent,gu2023two,zhu2023all}. To reflect the particular contribution of the technique and the problem that is addressed, questions are formulated distinctively in some cases: ``Choose an image that best and most realistically reflects the style" \cite{sanakoyeu2018style}, 
or ``Identify the patch to be real artwork from a stylized image" \cite{kotovenko2019content}. 

Measuring the ``realism" of fake paintings (stylizations) in comparison to human-created paintings has been in general a practice used in user surveys.
Studies ask participants if the shown image is a real artwork or not, reporting the results as ``Deception score" \cite{chen2021artistic,wang2022aesust}. A similar question (``Select the real artwork from a pair of a real artwork and a stylized image") was used by Huang~\etal \cite{huang2023quantart}. Zhang~\etal \cite{zhang2023unified} designed a user study in which participants were shown 10 artworks of which 2-4 were synthetically generated by the same method (one of the methods in comparison). The participants were asked to select the synthetically generated images, with precision and recall reported for each method. 

Hu~\etal \cite{hu2020aesthetic} asked the participants to vote for three results that are most similar to the style image in terms of color, texture, and overall feeling. 
Another practice is to use 3 or 4 questions per task whereby the user is asked to answer each question individually, with each question considering a different aspect of the stylization, such as content preservation/integrity, style quality/level, and overall quality. In rarer cases, user studies collect ratings for the aforementioned aspects \cite{huo2021manifold} or they ask the users to rank the displayed images in order from ``Best" to ``Worst" \cite{johnson2016perceptual}. 

\subsubsection{Photorealistic \& Video NST User Studies}
The user surveys in Photorealistic Image NST seem to follow a more standardized procedure. Put forward by the work of Luan~\etal \cite{Luan2017deep}, who also provided the most predominantly used test dataset, a study to evaluate the results of photorealistic approaches is two-fold. Firstly, the authors attempt to evaluate the realism of the results by asking the participants to rate a stylized image from ``Definitely photorealistic" to ``Definitely not photorealistic". Secondly, the participants are asked to choose the output image that is more similar to the presented style image. Although not identical, similar task presentation and question phrasing are followed by subsequent Photorealistic NST methods \cite{mechrez2017photorealistic,li2018closed,yoo2019photorealistic}. An~\etal \cite{an2020ultrafast} opted for a simpler approach, as the participants were only asked to choose the best image in terms of less artifact, less distortion and more details.

The user studies in Video NST generally follow the same design as in Image NST, with some obvious differences. First, instead of images, the participants are shown content and stylized videos. Additionally, the question formulation is revised to encompass a consideration for temporal coherence. More precisely, in addition to the resemblance to the style image \cite{ruder2018artistic,gao2018reconet,gao2020fast}, participants are asked to select the synthesized video that is the most temporally stable \cite{li2019learning,gao2018reconet,wang2020consistent,deng2021arbitrary,wu2022ccpl} and which avoids visual artifacts and flickering effects \cite{huang2017real,ruder2018artistic,gao2020fast}.

\subsubsection{Other considerations}
For both Image and Video NST, a multitude of factors regarding the design and reporting of user studies exist that can potentially influence the responses of the participants and the concluded outcomes \cite{bylinskii2023towards}. Infrequently, the authors undertake measures to facilitate the attainment of more reliable results. For example, Mechrez~\etal \cite{mechrez2017photorealistic}, when testing for realism, validated the responses of the participants by asking them to rate the original (realistic) image; if an average rating was less than a threshold, they excluded the responses of the respective participants to filter out poor-quality data. Similarly, Xu~\etal \cite{xu2018learning} ruled out untrustworthy results labelled too soon.

Bylinskii~\etal \cite{bylinskii2023towards} provide a detailed and thorough evaluation of the current trends regarding the design and reporting of user studies in computer graphics and vision, combined with robust recommendations on how to ameliorate inconsistencies and misleading results. Other factors that researchers might need to take into account are:
\begin{itemize}
    \item The means used to conduct the survey: Online questionnaire or face-to-face in a controlled lab environment?
    \item The image quality and resolution of the presented images.
    \item The screen specifications or the quality of the screen each participant has viewed to complete the survey.
    \item The exact placement of the images on the screen: Are all the images placed side-by-side, or are there images placed below or on top of others? Where are the content and style images placed relative to the results?
\end{itemize}

Given the discrepancies in the design of user studies in NST, it is important for future studies to establish a standardized human evaluation approach. This would allow for more accurate and reliable comparisons between different NST methods and their performance. Additionally, it would provide a clearer understanding of the potential of NST in various applications. Further analysis and recommendations regarding the user studies in NST are provided in Section~\ref{sec:analysis}.

\subsection{Quantitative Evaluation Metrics}

Depending on the problem a method attempts to solve, only a small selection of quantitative metrics are employed for evaluative comparisons (Figure~\ref{fig:quantitative_metrics_utilization}). We classify the metrics into four distinct categories depending on the particular aspect of the stylization performance they quantify:
\begin{itemize}
    \item \textbf{Perceptual Metrics}: Metrics that mostly focus on the content and structure preservation performance.
    \item \textbf{Stylization Performance}: Metrics that assess how well the stylized image resembles the style image.
    \item \textbf{Video}: Metrics that gauge temporal coherence and video stability performance.
    \item \textbf{Efficiency}: Metrics for benchmarking characteristics regarding the system's performance.
\end{itemize}

\subsubsection{Perceptual Metrics}
\label{sec:perceptual_metrics}

To assess the content preservation performance of NST models, various widely used perceptual metrics are utilized. Typically, the perceptual metrics in computer vision research measure the similarity between two images. In NST, these metrics are computed between the original content images and the stylized results. Most of the metrics do not simply compare pixel values; instead, the majority of those take into account higher-level features such as edges, textures, and colors.

One of the most frequently adopted metrics is Structural Similarity Index (SSIM) \cite{wang2004image} defined as:
\begin{equation}
    SSIM(x,y) = \frac{(2\mu_{x}\mu_{y} + c_1)(2\sigma_{xy} + c_2)}{(\mu_{x}^{2}+\mu_{y}^{2}+c_1)(\sigma_{x}^{2}+\sigma_{y}^{2}+c_2)}
\end{equation}
where $x$ and $y$ are the two images being compared, $\mu_{x}$ and $\mu_{y}$ are the pixel sample means of the two images, $\sigma_{x}^{2}$ and $\sigma_{y}^{2}$ are the variance of $x$ and $y$ respectively, $\sigma_{xy}$ is the covariance of $x$ and $y$, and $c_1, c_2$ are constants used to prevent the denominator from becoming zero. SSIM is based on the degradation of structural information and for its computation it takes into account luminance, contrast and structure \cite{wang2004image}. 

With the advent of deep learning, other perceptual metrics have been proposed that are considered to be more compatible with the human visual system and capable of more effectively capturing the perceptual characteristics of images. The Learned Perceptual Similarity metric (LPIPS) works by extracting and comparing features from pre-trained neural networks, shown to model low-level perceptual similarity particularly well \cite{zhang2018unreasonable}. Another commonly used metric is Content Error, which computes the mean square error of feature activations between the stylization and the content image, similar to the content loss function employed during the training of most NST methods.

To analyze the effectiveness of the NST systems in retaining content information, An~\etal \cite{an2021artflow} proposed the \textit{Content Leak} phenomenon. To gauge the amount of content leak, a content-style pair is used to perform a stylization. Then, using the stylized result as the new content image, the style transfer process is performed repeatedly multiple times. It is then easy to notice whether content information is lost/retained in the resulting outputs.

One of the three quantifiable metrics suggested by Wang~\etal \cite{wang2021evaluate} is Content Fidelity (CF) which can measure the faithfulness of the stylized result ($y$) to the original content image ($x$), at multiple scales. It utilizes cosine similarity to measure the differences in deep feature activations ($f_{l}$):
\begin{equation}
    CF(x,y) = \frac{1}{N} \sum_{l=1}^{N} \frac{f_{l}(x) \cdot f_{l}(y)}{\|f_{l}(x)\| \cdot  \|f_{l}(y)\|}
\end{equation}
where $N$ is the number of different layers.

Approaches that are more focused on retaining global structure and depth information \cite{liu2017depth,cheng2019structure,ioannou2022depth} further use histogram and histogram-based methods that consider intensity and tone information within the image. The histogram is a depiction of the distribution of pixel values in the image. It is useful for detecting tonal and color differences between the content image and the stylized result. Similarly, the image hash algorithms \cite{buchner_2021} -- average hash (aHash) and difference hash (dHash) -- analyze the image structure on luminance and they are suitable for identifying similarities in the input images.

These approaches also attempt to quantify the depth and content preservation capabilities of their model using depth, edge, and saliency map comparisons. This information is inferred using state-of-the-art approaches (e.g., depth prediction \cite{chen2016single,Ranftl2020}; edge detection \cite{xie2015holistically,liu2017richer}). 
Saliency detection, which is considered an instance of image segmentation, can also be computed with models such as the one by Jiang~\etal \cite{jiang2013salient}.
Comparisons between the derived depth/edge/saliency map of the content image and the corresponding map of the stylization image are performed using standard similarity measures, such as mean square error or SSIM.

\subsubsection{Stylization Performance} 

Assessing how well the stylized result resembles the reference style images is a challenging task, and maybe a subjective one. Although multiple metrics have been proposed for the evaluation of the stylization performance of NST methods, no metric is currently used as the gold standard. This group of metrics compares the outputs of the NST methods with the corresponding style images that are used to generate them.

Analogous to the Content Rrror used to measure content preservation, Style Error is also employed. This makes use of features extracted from pre-trained models and is defined identically to the style loss used in training -- the mean square error of Gram matrices between the feature activations of the style image and the stylized result.

Kotovenko~\etal \cite{kotovenko2019content} proposed a method for content and style disentanglement. As part of their evaluation, they employed a statistical distance metric to measure how well their system is capable of covering the style distribution it attempts to reproduce. Using a trained network on painting classification, they extract activations on real artworks to define the true style distribution $\mathbb{P}_{s}^{art}$, and activations from the stylizations derived from the proposed style transfer model to approximate $\mathbb{P}_{s}^{stylized}$. Then the Kullback-Leibler Divergence is computed:
\begin{equation}
    D_{KL}(\mathbb{P}_{s}^{stylized}, \mathbb{P}_{s}^{art})
\end{equation}
depicting how well the style distribution is represented.

Style Transfer Deception rate, suggested by Sanakoyeu~\etal \cite{sanakoyeu2018style}, is based on a similar idea. A \textit{VGG-16} network trained to classify artists from the Wikiart dataset, is used to classify multiple stylizations. The deception rate is then calculated as the fraction of synthesized images classified to belong to the artist whose style they attempted to replicate.

In addition to Content Fidelity (CF) defined in Section~\ref{sec:perceptual_metrics}, Wang~\etal \cite{wang2021evaluate} also proposed two effects to assess stylization performance. The Global Effects (GE) metric combines Global colors (GC) that measure color histogram differences ($hist_c$):
\begin{equation}
    GC(y,s) = \frac{1}{3} \sum_{c=1}^{3} \frac{hist_c(y) \cdot hist_c(s)}{\|hist_c(y)\| \cdot \|hist_c(s)\|}
\end{equation}
where $y$ and $s$ are the stylized result and style image respectively; and Holistic Textures (HT) that is similar to the Style Error measurement that uses Gram matrices $G_l$:
\begin{equation}
    HT(y,s) = \frac{1}{N} \sum_{l=1}^{N} \frac{G_l(y) \cdot G_l(s)}{\|G_l(y)\| \cdot \|G_l(s)\|}
\end{equation}
Then Global Effects is defined as: $GE(y,s) = \frac{1}{2} (GC(y,s) + HT(y,s))$. A different factor was also suggested that measures the quality of local style patterns. The Local Patterns (LP) factor, defined as $LP(y,s) = \frac{1}{2} (LP_1(y,s) + LP_2(y,s))$ where $LP_1$ measures differences of local patterns counterparts directly ($\phi^l_i(y)$ and $\phi^l_j(s)$ are used to denote the neural patches for multi-scale features):

\begin{equation}
    LP_1(y,s) = \frac{1}{Z} \sum_{l=1}^{N} \sum_{i=1}^{n_y} \frac{\phi_i^l(y) \cdot \phi^l_{CM(i)}(s)}{\|\phi^l_i(y)\| \cdot \|\phi^l_{CM(i)}(s)\|}
\end{equation}
where $CM(i) := \arg\max_{j=1,...,n_s} \frac{\phi^l_j(y) \cdot \phi^l_j(s)}{\|\phi^l_j(y)\| \cdot \|\phi^l_j(s)\|}$, and $LP_2$ compares the diversity of the pattern categories:

\begin{equation}
    LP_2(y,s) = \frac{1}{N} \sum_{l=1}^{N} \frac{t_{CM}^l}{t_s^l}
\end{equation}
where \( t^l_{cm} \) and \( t^l_s \) are the numbers of \( \phi^l_{CM(i)}(s) \) and \( \phi^l_j(s) \), respectively.

Ruta~\etal \cite{chen2023learning} uses Chamfer distance to measure color consistency. Here, Chamfer distance is used to measure dissimilarities between the stylization and the reference style image in image color space. Formally, Chamfer distance calculates the minimum distance between each point in one set $X$ (of pixels) and the nearest point in the other set $Y$ (of pixels), and sums up these distances: 

\begin{equation}
    CD(X, Y) = \sum_{x \in X} \min_{y \in Y} ||x-y||_{2}^{2} + \sum_{y \in Y} \min_{x \in X} ||x-y||_{2}^{2}
\end{equation}

To assess color differences, Gu~\etal \cite{gu2023two} employed the color loss proposed in \cite{ignatov2017dslr}. This computes the Euclidean distance between the (Gaussian) blurred versions of the style image ($s_b$) and the stylized result ($y_b$):
\begin{equation}
    \mathcal{L}_{\text{color}}(s,y) = \|s_b - y_b\|_2^2
\end{equation}

Another metric for measuring the deviation between the distribution of deep features of generated images and that of real images is the Fréchet Inception Rate (FID) \cite{heusel2017gans}, commonly used for the evaluation of GAN methods. This can be suitable for evaluating stylizations by adjusting it to work for a single image and internal patch statistics (SIFID) \cite{shaham2019singan}. This can thus be utilized to measure style consistency between the style images and the stylized results \cite{ruta2023neat}. 

More recently, an enhancement to the FID metric has been proposed that attempts to evaluate both content preservation and style matching \cite{wright2022artfid}. The \textit{ArtFID} metric uses the Inception network trained on a large-scale art classification dataset and computes feature distribution differences between image features extracted from the stylizations and image features extracted from the style images. The distance between the two feature distributions is calculated using the Fréchet distance. \textit{ArtFID} is thus formulated as:

\begin{equation}
    \begin{aligned}
    \text{ArtFID}(X_g, X_c, X_s) = \left(1 + \frac{1}{N} \sum_{i=1}^{N} d(X^{(i)}_c, X^{(i)}_g)\right) \\ \cdot 
     \left( 1 + \text{FID}(X_s, X_g)\right)
    \end{aligned}
\end{equation}
where $X_g$, $X_c$, and $X_s$ are the stylized images, the content images and the style images respectively, $d(X^{(i)}_c, X^{(i)}_g)$ is measured using the LPIPS metric to account for content preservation and FID is defined as: \begin{equation}
    \text{FID}(X_s, X_g) = \|\mu_s - \mu_g\|_2^2 + \text{Tr}(\Sigma_s + \Sigma_g - 2(\Sigma_s\Sigma_g)^{\frac{1}{2}})
\end{equation}
where $(\mu_s,\Sigma_s)$ and $(\mu_g,\Sigma_g)$ correspond to the mean and covariance of the extracted Inception features of the style images $X_{s}$ and stylized images $X_{g}$ respectively, and $Tr$ refers to the trace linear algebra operation. \textit{ArtFID} is shown to be compatible with human judgement \cite{wright2022artfid}.

Another metric that is also shown to strongly coincide with human judgment is \textit{ArtScore} \cite{chen2023learning}. This measures how well a synthesized image resembles an authentic artwork. The framework proposed by Chen~\etal \cite{chen2023learning} applies transfer learning using StyleGAN models \cite{karras2020analyzing}, employs interpolation image generation and trains a neural network (\textit{ArtScore}) with ResNet-50 pre-trained on ImageNet as the backbone and with an effective learn-to-rank objective.

\subsubsection{Video} 

Except for the perceptual and style performance metrics, methods suitable for stylizing video sequences are required to quantify the effectiveness of their system in achieving temporal stability. As most of the approaches work in real-time -- generating a fast stylization for a single frame through a forward pass -- the derived stylizations are prone to flickering effects and undesired instabilities. The most reliable way to quantify temporal consistency is the warping error which calculates the difference between a warped next frame and a ground truth next frame \cite{lai2018learning}. 
\begin{equation}
    WarpingError = \sqrt{\frac{1}{T-1}\sum_{t=1}^{T} || O_{t} - W_{t}(O_{t-1}) ||^{2}} 
\end{equation}
where $O_{t}$ and $O_{t-1}$ are the corresponding output stylized frames of the current and previous input frames, $W_{t}$ is the ground truth optical flow, and $T$ is the number of time steps or the number of frames in a sequence. Methods that do not evaluate videos with available optic flow masks might compute those using state-of-the-art approaches, such as \textit{FlowNet} \cite{dosovitskiy2015flownet}. Some approaches may also include the ground truth occlusion mask in the computation to account only for traceable pixels.

More straightforward approaches to measuring temporal stability include calculating the LPIPS (Section~\ref{sec:perceptual_metrics}) between adjacent frames or the mean and variance of the subsequent video frames. LPIPS provides a way to compute the average perceptual distances between adjacent frames. Deng~\etal \cite{deng2021arbitrary} chose to quantify temporal smoothness by defining $Diff_{F(t)} = ||F_{t} - F_{t-1}||$ and calculating the mean and variance of $Diff_{F(t)}$, where $F_{t}$ and $F_{t-1}$ are two adjacent frames of a T-frame rendered video. Similarly, Lu and Wang \cite{lu2022universal} compute the average pixel distances using $D^{*}$ defined as:
\begin{equation}
    D^{*} = \frac{1}{T-1}\sum_{t=0}^{T-2} ||F_{t+1} - F_{t}||_{2}
\end{equation}

To quantify the temporal stability performance of their Photorealistic Video style transfer method, Xia~\etal \cite{xia2021real} employed a Temporal Change Consistency metric (TCC) \cite{zhang2019exploiting}, which is based on depth maps:
\begin{equation}
    TCC(D,G) = \frac{\sum_{i=1}^{n-1} SSIM(abs(d^{i} - d^{i+1}), abs(g^{i} - g^{i+1}))}{n-1}
\end{equation}
where $D = (d^{1}, d^{2}, ..., d^{n}) $, and $G = (g^{1}, g^{2}, ..., g^{n}) $ are the estimated depth maps of the stylized frames, and the corresponding ground truth depth maps, respectively.  

\subsubsection{Efficiency} 

Another essential aspect of the quantification of the performance of a stylization algorithm is efficiency. This basically accounts for speed, memory and control. NST algorithms compare the time their model needs to output a stylized image through a forward pass given an input content image (e.g., \cite{Risser2017stable,johnson2016perceptual,deng2022stytr2,liu2021adaattn,an2021artflow}). Occasionally, the memory a trained model requires is compared to state-of-the-art models, especially when approaches are focused on improving the stylization network and its composition (e.g., \cite{li2016precomputed,sanakoyeu2018style,zhang2018multi,shen2018neural}). Another consideration has to do with user control (e.g., allowing users to stylize specific areas of an image differently). Unless there is an obvious difference in the number of styles a model can reproduce (e.g., Per-style-per-model versus Arbitrary-style-per-model), the number of styles a method is capable of emulating is also compared \cite{gao2020fast}.

\subsection{Ablation Studies}
\label{sec:Ablation Studies}

Inspired by the field of neuroscience, where ablation studies have been used to unveil and analyze more precisely the structure and organization of the human brain, artificial intelligence, machine learning and subsequently computer vision research has resorted to the concept of removing (or substituting) specific components from artificial neural networks to effectively analyze their behaviour \cite{meyes2019ablation}. NST systems have also used this idea, which is capable of deriving useful insights and understanding, resulting in an in-depth evaluation of the performance of the proposed models. Ablations cannot be explicitly regarded as Qualitative or Quantitative Evaluation techniques, as they can be used to produce both types of data -- for the examination of particular components or aspects of an approach, qualitative results (e.g., stylized images/videos) can be provided, or the computational metrics can be re-run for a quantitative assessment. 

\begin{figure}[htb]
    \centering
    \includegraphics[width=\linewidth]{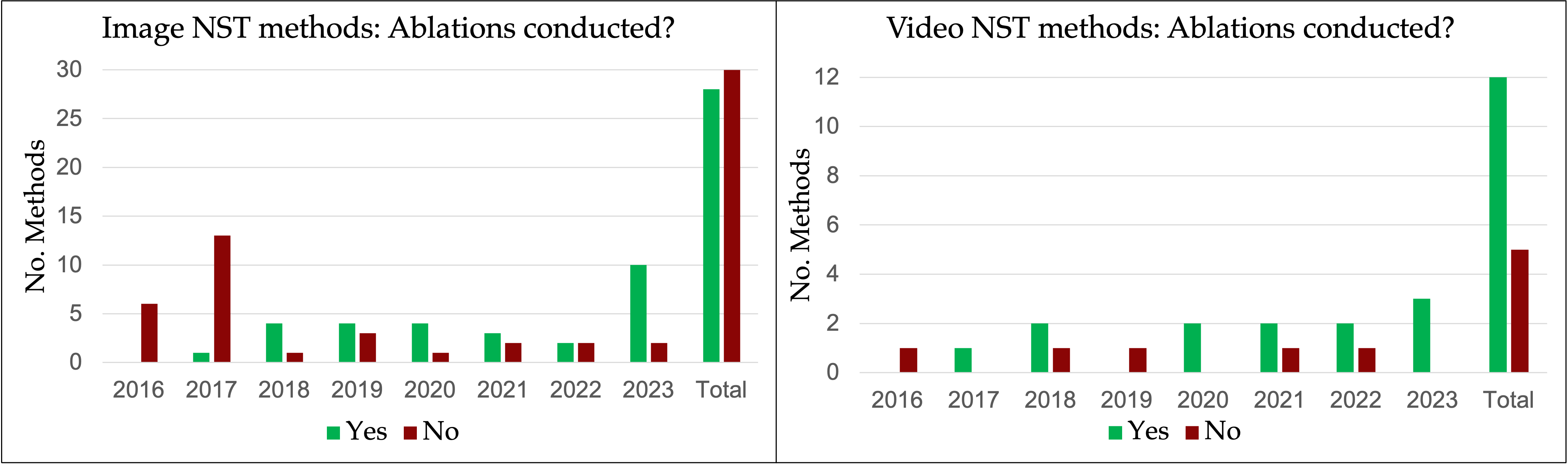}
    \caption[Ablation Studies in Image and Video NST]{The number of Image and Video NST studies including Ablation Studies. The graph depicts the reviewed papers of the NST literature.}
     \label{fig:ablations}
\end{figure}

Figure~\ref{fig:ablations} provides an overview of the involvement of ablation studies in Image and Video NST evaluation approaches. 
More Video NST studies include ablations than not, based on the papers reviewed in this work (Table~\ref{tbl7}). 

A common practice is the study of the effect of the different losses the NST methods employ during training \cite{sanakoyeu2018style,kotovenko2019content,gu2018arbitrary,liu2021adaattn,park2019arbitrary,luo2022consistent}. Yet, as the design and implementation of NST models can be widely diverse, the ablation studies employed assess different aspects in different methods. For example, Deng~\etal \cite{deng2021arbitrary} performed ablations to analyze the effect of the content-aware positional encoding (CAPE) method proposed, whereas Liu~\etal \cite{liu2021adaattn} provided qualitative results to verify the effectiveness of the shallow feature used in their suggested method (\textit{AdaAttN}). Ablation studies can be crucial for enhancing interpretability and understanding. 
\subsection{Experiments}
\label{sec:experiments}

\begin{figure*}[htb]
    \centering
    \includegraphics[width=\linewidth]{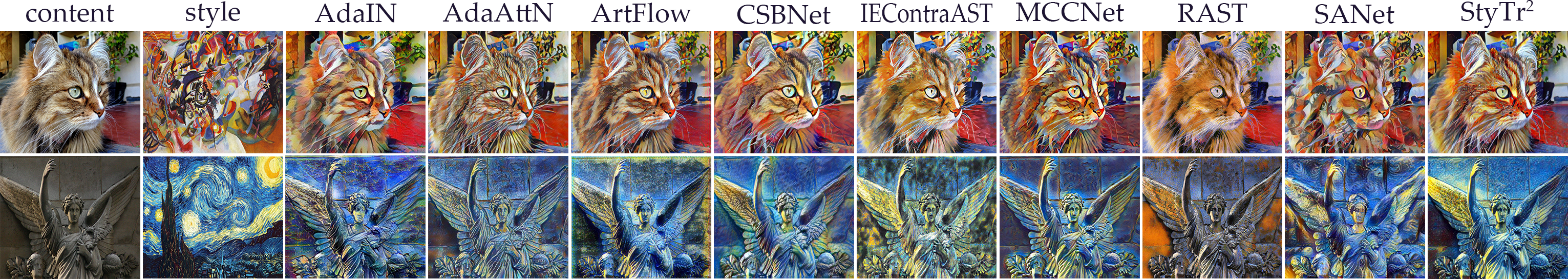}
    \caption[Results of state-of-the-art NST methods]{Results of state-of-the-art NST approaches. More results are provided in the supplementary material.}
     \label{fig:experiments}
\end{figure*}

To show an example of a current evaluation procedure and reveal some of the potential issues that arise, in this section, we experimentally compare nine state-of-the-art methods: \textit{AdaIN}~\cite{huang2017arbitrary}, \textit{AdaAttN}~\cite{liu2021adaattn},  \textit{ArtFlow}~\cite{an2021artflow}, \textit{CSBNet}~\cite{lu2022universal}, \textit{IEContraAST}~\cite{chen2021artistic}, \textit{MCCNet}~\cite{deng2021arbitrary}, \textit{RAST}~\cite{ma2023rast}, \textit{SANet}~\cite{park2019arbitrary}, and
\textit{StyTr\textsuperscript{2}}~\cite{deng2022stytr2}. We utilize the dataset proposed by Mould and Rosin \cite{mould2017developing} comprised of 20 content images, and 10 style images that are frequently employed by NST methods and which encompass a diverse range of stylistic attributes (e.g., genre, color, texture). Example results are shown in Figure~\ref{fig:experiments}. Although the techniques differ, the stylizations produced are quite similar. Based on the visual side-by-side comparisons alone, it is difficult to quantify the performance of the methods and to estimate which approach preserves the content information better, or which approach generates results that capture the style image more accurately.

Table~\ref{tab:experiments} gives quantitative comparisons for the state-of-the-art methods. Three different computational metrics are used to capture the content preservation performance while two different computational metrics are employed to quantify style resemblance. \textit{RAST} generates consistently more content-aware results that preserve the global structure of the input content images. However, for the rest of the methods, there is a lot of variability in their performance across the different metrics. For example, \textit{ArtFlow} performs better than \textit{StyTr\textsuperscript{2}} in Content Error ($\mathcal{L}_{c}$), but it falls behind in SSIM. Similarly, when quantifying the style performance of the algorithms, \textit{IEContraAST} achieves better results than \textit{AdaAttN} when considering SIFID, but \textit{AdaAttN} outperforms \textit{IEContraAST} when measuring Style Error ($\mathcal{L}_{s}$). Additionally, while \textit{AdaAttN} performs competently to the rest of the approaches on the Style Error metric, its efficacy drops significantly on the SIFID metric.

\begin{table}[htb]
\setlength\extrarowheight{1pt}
\centering
    \caption{Quantitative comparisons between state-of-the-art NST approaches. $\mathcal{L}_{c}$ and $\mathcal{L}_{s}$ account for Content and Style Loss (Error) respectively. The best results are in bold and the second-best results are marked with an underline.}
    \label{tab:experiments}

    \begin{tabular}{| l | c | c | c | c | c |}
        \hline
             Method & $\mathcal{L}_{c} \downarrow$ & SSIM $\uparrow$ & LPIPS $\downarrow$ & $\mathcal{L}_{s} \downarrow$ & SIFID $\downarrow$ \\ 
       \hline
       
        \textbf{AdaIN} & 1.6170 & 0.2588 & 0.5174 & \textbf{0.1023} & \textbf{0.6223} \\ \hline
        
        \textbf{AdaAttN} & 1.2542 & 0.4705  & 0.4766 & \underline{0.1049} & 18.3407 \\ \hline
        
         \textbf{ArtFlow} & \underline{0.9575} & 0.4547 & 0.4824 & 0.1079 & 0.9657 \\ \hline

        \textbf{CSBNet} & 1.2494 & 0.3601  & 0.4780 & 0.1141 & 5.0953 \\ \hline

        \textbf{IEContraAST} & 1.3530 & 0.4065 & \underline{0.4439} & 0.1064 & \underline{0.8600} \\ \hline

        \textbf{MCCNet} & 1.1193  & 0.4547  & 0.4714 & 0.1091 & 2.3755 \\ \hline

        \textbf{RAST} & \textbf{0.9272} & \textbf{0.5383} & \textbf{0.3101} & 0.1113 & 1.7580 \\ \hline

        \textbf{SANet} & 1.4260 & 0.3096 & 0.5373 & 0.1073 & 1.1592 \\ \hline
        
         \textbf{StyTr\textsuperscript{2}} & 1.0415 & \underline{0.4882} & 0.4693 & 0.1073 & 1.1089 \\ \hline     
                
    \end{tabular}
    
\end{table}

Although quantitative computational metrics can offer valuable insights into the efficiency of the algorithms, their application varies across NST methods, leading to inconsistencies in evaluation. The next section provides an in-depth analysis of the considerations arising from the utilization of the various evaluation techniques and proposes ways to alleviate them.

\section{Analysis and Recommendations}
\label{sec:analysis}

\subsection{Benchmark Datasets}
\label{sec:benchmark-datasets}

As discussed in Section~\ref{sec:3-datasets}, for Video NST there is some agreement as to what data is used in the evaluation. Similarly, in the domain of photorealistic Image NST, there is some consensus on the dataset employed. However, there is no consensus within the domain of artistic Image NST. 

Only a few approaches have focused on addressing the lack of a benchmark dataset in NST research. The method of Mould and Rosin \cite{mould2017developing} suggested a set of content images that fulfil a set of criteria and can potentially be utilized as a benchmark for evaluation. However, they don't consider style images and the potential characteristics that a benchmark style image set should include. The recent approach of Ruta~\etal \cite{ruta2023neat} proposed a large-scale dataset for style transfer, containing both content and style images. The suggested \textit{BBST-4M} dataset categorizes images into style and content based on how stylistic they are. While this dataset holds significant potential, compared to the work of Mould and Rosin \cite{mould2017developing}, it falls short in considering the distinct characteristics and attributes that should be encompassed by both the content image set and the style image set. The difficulty in composing a suitable evaluation dataset can be attributed to the existence of various types of evaluation and to the diverse intentions and goals of the proposed stylization solutions.


A summary of the research questions to address when compiling a sufficient evaluation dataset is:
\begin{itemize}
    \item What are the properties or attributes the content image set and style image set should include? In accordance with this, does the dataset distribution cover a sufficient amount of the identified image characteristics?
    
    \item Can an NST approach utilize the dataset effectively to demonstrate a new solution to an existing or new problem? In accordance with this, does the evaluation benchmark consider a sufficient amount of possible use cases, covering a wide range of potential problems and applicability of NST? 

     \item Is the evaluation benchmark suitable for both Qualitative and Quantitative evaluation, including Human Evaluation Studies? Or, should the dataset be divided into three partitions, with each image set comprising the relevant images based on the specific type of evaluation? What is the optimal size of the dataset?
    
\end{itemize}

\subsubsection{Content \& Style Benchmark datasets}

Mould and Rosin \cite{mould2017developing} discussed a range of policies for selecting the proposed image set. Although they only provide a small set of content images, the principles with which the set of images is composed are worth revisiting. Their suggested list of image properties includes colorfulness, complexity, contrast, sharpness, lineness, mean, standard deviation and noise. For these, a numerical value can be derived using computational methods, thus enabling an evaluation of how broad the distribution of a set of images is and the range of properties it covers. A list of image properties, explicitly suited for designing the benchmark set of content images is also provided. This includes variation in scale, fine detail, variation in texture, regular structure, vivid and varied colors, muted colors, thin features, human faces, and more. Independently from the discussion on the size of the evaluation benchmark and the number of content images and style images that should be utilized in each stage of the evaluation process, the set of 20 images (\textit{NPGeneral}) presented by Mould and Rosin \cite{mould2017developing} could serve as a promising initial foundation and offer valuable insights for the design of a comprehensive evaluation benchmark dataset.

Although there have not been any efforts to establish a benchmark video dataset, the literature suggests a convergence on utilizing videos from the MPI Sintel dataset for evaluation (Section~\ref{sec:video-nst-data}). This dataset is useful as it also contains ground truth optical flows for measuring the warping error and gauging the temporal coherence performance of the methods in comparison. Nevertheless, its synthetic nature does not allow for a comprehensive and detailed examination of the results, as the performance might degrade when tested on real-world videos \cite{man2022review}. We argue that a similar effort to the approach of Mould and Rosin \cite{mould2017developing} should be made in order to develop a video dataset that encompasses a wide variety of characteristics. Videos could be retrieved from multiple sources, combining both synthetic and real-world scenes. Establishing such a benchmark dataset for Video NST approaches could result in more robust comparisons and insightful evaluations. 

Large-scale art datasets (e.g., Wikiart) can greatly facilitate the creation of a well-suited set of style images. An evident evaluation dataset design decision would be to compose the style images set by incorporating a diverse range of artworks encompassing various artistic genres, e.g., as defined in Wikiart. It would therefore be necessary to provide an analysis and discussion regarding the definition of an artistic genre and its interrelation with style. As defined in \textit{Tate} \cite{tate2023}, 
genres are types of painting codified in the 17th century as history, portrait, genre painting (scenes of everyday life), landscape and still life. Developed in European culture, the genre system is not particularly relevant to contemporary art but is a system that can be used to divide artworks according to depicted themes and objects. The Wikiart dataset includes 68 different genres of art (not all of them consider visual 2D artworks), e.g., ``abstract", ``graffiti". 
This visual art encyclopedia distinguishes between genres and styles, defining style as the distinctive visual elements of the artwork, its techniques and methods, usually corresponding with an art movement (e.g. ``Cubism"). 
Arguably, a comprehensive benchmark style image set should include artworks spanning a broad spectrum of artistic genres and styles. However, a range of image properties (e.g., colorfulness, complexity, contrast) should be also covered.

\subsubsection{Different use cases \& Different types of evaluation}
Since the seminal work of Gatys~\etal \cite{gatys2016image}, style transfer approaches have emerged that consider different aspects and introduce different perspectives. Proposed methods suggest an aesthetic standpoint, decomposing style into color and texture \cite{hu2020aesthetic}, attempt to disentangle content and style \cite{kotovenko2019content}, introduce geometric warps \cite{liu2021learning}, or utilize depth information to retain global structure and depth effect \cite{liu2017depth,cheng2019structure,ioannou2022depth}. When considering the benchmark set of content images, it is essential to incorporate a wide variety of examples, in order to accommodate all the different use cases. A further examination of the data utilized in all the different kinds of approaches would allow for an informed design of the content image space, adding to the list of image properties and attributes.

Another crucial consideration regarding the design of an evaluation benchmark is the existence of different types of evaluation. The number of subjective visual side-by-side comparisons that can be presented is limited and there are constraints to the amount of questions recruited human participants are able to respond to in the setting of a user study. For quantitative metrics, an important consideration that should be taken into account is that the larger the dataset, the more reliable the quantitative results would be.

\subsubsection{Suggested policies}
Based on these considerations, policies regarding the design of benchmark sets of images for use in the different evaluation practices can be formulated. For the visual comparisons accompanied by the subjective judgements of the authors in an NST paper, it would be beneficial for the NST community if the authors included a universally agreed-upon small set of content and style images in the results. The authors could complement the suggested benchmark images with their own images, but including some common images in all studies would allow for fairer and more insightful comparisons. For the Human Evaluation Studies, another set of content and style images can be established. This can be larger than the set of images presented in the paper, as a larger sample can provide more statistical significance to the results of a user study. Further discussion regarding the data and best practices are discussed in the next sections.

Except for the image properties that the content and style image set should satisfy, the size of the benchmark test dataset for the quantitative evaluation is another significant aspect. The larger the benchmark evaluation dataset for the computation of the quantitative metrics, the more representative and accurate the resulting comparisons would be. Theoretically, NST research could follow the standard practices in computer vision research using deep learning approaches. A split (e.g., 80\% - 20\%) on the training dataset could accommodate a test set. However, the most commonly used datasets (Wikiart and MS COCO), might not sufficiently satisfy the aforementioned requirements for a good evaluation benchmark. 

It would be essential to consider the size of the evaluation benchmark in relation to the average training set. A recommended guideline could be to ensure that the evaluation benchmark is at least 10\% of the size of the average training set. This ensures that the evaluation dataset is representative and provides sufficient coverage of different content and style variations. An appropriate dataset, for example, would be composed of a combination of content and styles that when multiplied yields a number of stylized results, the total count of which could be proportionate to the magnitude of the average training set utilized. If we consider MS COCO as the most common dataset, the total number of stylizations should approximately be 8000 (10\% of MS COCO). Another important design decision is the number of samples in the content set and the styles set. If \textit{NPGeneral} \cite{mould2017developing} is used as the benchmark content image set, the size of the benchmark styles set should be 400. 
As it is more crucial to evaluate how well a model generalizes in reproducing the artistic effect of a variety of artworks, the style images set could be designed first to be the smallest possible that can encompass all the different styles. Then, the content image set can be fixed accordingly. Sanakoyeu~\etal used 18 styles and produced 300 stylizations per style to measure the deception rate. An image set of approximately 50 styles, combined with a content image set of size 100, would result in 5000 stylizations that could suffice for the requirements of quantitative evaluation. A more thorough discussion of Quantitative evaluation best practices is provided in Section~\ref{sec:analysis-quantitative}.

Currently, there does not exist a benchmark evaluation dataset specifically tailored for NST methods, and no substantial efforts have been made to compile a large-scale dataset that can fit the evaluation requirements of NST research. Hence, it is essential to emphasize the significance of providing all the relevant and necessary information regarding the data that is utilized during the evaluation process, so that any presented results are easily repeatable and reproducible. Initiatives (e.g., \textit{Plan S: \url{https://www.coalition-s.org/}}) are already in place which seek to establish an open-research practice among the scientific community. High-impact journals 
have committed to making published research open-access  \cite{acm2023}. We encourage the NST community to also adopt these disciplines, releasing both the data and code of any novel approach. Thus, it is of great significance for the field of NST that the evaluation data is not only published but also made openly accessible. By doing so, the validity and reliability of the research findings could be significantly enhanced. This practice promotes transparency and would allow for independent verification and further exploration of the results.

\subsection{Qualitative Evaluation}

Despite relying on subjective judgements, qualitative evaluation can provide useful insights regarding the effectiveness of a proposed method. By presenting visual results side-by-side with state-of-the-art methods (e.g., Figure~\ref{fig:experiments}), the differences and novelty in a proposed technique's results become evident. The inclusion of additional zoomed-in cut-outs can effectively emphasize significant distinctions and provide enhanced visibility of the improvements made.

As described in the previous section, it could be feasible and potentially valuable if a universally agreed-upon image dataset is used for the qualitative evaluations. This could consist of a reasonably small amount of content and style images, resulting in a small number of stylizations that are derived using the same combinations of content and style in all the NST papers. Definitely, the size of this set of images would probably not be adequate to capture and highlight the effectiveness of the proposed method, therefore, the authors could complement these stylizations with more images that they chose that could be useful for the readers. However, as it would be beneficial for the field if each method included the same comparisons, a website maintained by the authors of each paper or the corresponding supplementary material section could be used to accommodate a substantial amount of image comparisons.

In the supplementary material, we include an illustration the most commonly used content and style images. Considering that the existing literature predominantly consists of qualitative evaluations relying on this particular image set, we argue that the small benchmark dataset, encompassing a limited number of content and style images, can be constructed using images from this set. This will enable more consistent literature and allow NST researchers to draw visual comparisons with chronologically older methods.


\subsection{Human Evaluation Studies}

For the Human Evaluation Studies, the dataset utilized to produce the results shown to the participants is not the only point of disagreement among the NST methods. Different formulations of the questions, different presentations of the comparisons, and varied numbers of participants are also among the aspects contributing to inconsistencies in conducting user studies in the field. For a comprehensive analysis and overview of the most effective practices in user studies for computer graphics research, readers are encouraged to refer to the tutorial by Malpica~\etal \cite{malpica2023tutorial}.

\subsubsection{Data in NST User Studies} 

As argued in Section~\ref{sec:benchmark-datasets}, it is possible to establish a benchmark evaluation dataset for user evaluation studies. Depending on the total number of questions presented, a set of content and style images can be designed to capture a wide variety of image characteristics. As already noted, the greater the number of questions posed, the more significant the results can be. For example, a user study of 50 questions can employ results synthesized from a benchmark dataset of 10 content and 5 style images. This is still a small dataset (50 images), and might not sufficiently capture the novelty and efficiency of a proposed method -- a subset of the dataset can be replaced with a set of images that the authors deem to be suitable. In any case, it is recommended that the authors make the user study datasets available, as this is not only good practice for the reproducibility of the results, but could also encourage subsequent studies to utilize the same dataset or build on it.

\subsubsection{Participant considerations}
\label{sec:participant-considerations}

Unlike other machine learning or computer vision tasks (e.g., object detection) that might require crowd-sourcing for evaluating the effectiveness of the results, evaluating results in NST research can be a subjective activity. Nevertheless, collecting and reporting supplementary data from participants regarding their experiences can significantly enrich our comprehension of the results and shed light on participants' behaviour. Such additional information -- demographics and task-related subjective data -- can provide valuable insights that enhance the overall understanding and interpretation of the findings \cite{cowley2022framework}. Importantly, the relation of the participants to the task and to the broader NST research or any art-adjacent field should be reported, as it can potentially influence their choices. The number of participants should be kept as high as possible, but without compromising the reliability and quality of the responses.

\subsubsection{Compared Methods} 

The abundance of NST methods (Tables~\ref{tbl1}-\ref{tbl7}) renders it impractical to directly compare any single approach to the entirety of available state-of-the-art techniques. Yet, it is important to compare with methods that attempt to solve a similar problem or view NST with a similar lens. For example, depth-aware methods are required to compare against state-of-the-art methods that utilize depth and aim for global structure and detail preservation. 
By including a comprehensive set of methods in the comparison, researchers can establish a more robust evaluation framework and provide stronger evidence to support their effectiveness in synthesising stylized visual imagery. 

\subsubsection{Presentation \& Question Formulation}

One of the most essential and critical aspects of a user evaluation study is the presentation of the questions, their exact formulation, and the format of the responses that are collected. As discussed in Section~\ref{sec:user-studies}, there is a variety of methods employed to present the questions and collect responses from the participants. Asking the participants to select their favorite stylization amongst all the stylizations displayed together is the most common approach. However, A/B testing and one-by-one presentation asking the participants to rate on a Likert scale are also employed. It is also worth noting that the Two-Alternative-Forced-Choice Task (2AFC) has been shown to be more precise when measuring aesthetic preference \cite{so2023measuring}. Nevertheless, it is difficult to reach a consensus as to which approach is most appropriate for each aspect of stylization (content preservation, style resemblance, overall aesthetics). Independent of the technique that is followed, we suggest the use of accompanying questions that allow the participants to justify their responses.  Collecting textual feedback for each question (or a majority) regarding the subjective judgement of the participants (combined with participant-related data, as mentioned in Section~\ref{sec:participant-considerations}), would allow for an improved understanding of human cognitive behaviour and the interpretation of the results. Asking the participants to justify their selection of one stylization over another would allow the users to point out details or highlight particular regions of the images where the differences are more obvious.

It is also important to acknowledge that the validity of any results is significantly impacted by the reliability of the recruited participants (Section~\ref{sec:participant-considerations}). The participants should be sampled representatively; ideally, the participants would have an interest in the NST software, and ultimately they could potentially use it \cite{bylinskii2023towards}. Nevertheless, this is hard to control, especially when utilizing crowdsourcing platforms, such as Amazon Mechanical Turk (AMT). It is possible, however, to filter out poor-quality data and avoid the potential carelessness of remote crowdworkers. An example is to remove unreliable data labelled too soon \cite{xu2018learning}. Both Bylinskii~\etal \cite{bylinskii2023towards} and Cowley~\etal \cite{cowley2022framework} propose the utilization of checks randomly inserted through the study. Embedding checks/questions where there is an objectively correct answer (e.g., produce obviously bad stylizations, maybe through the content leak phenomenon, as described in Section~\ref{sec:perceptual_metrics}) can allow detection of when participants are not paying careful attention, and help eliminate substandard responses. 

Another question presentation issue is the display and positioning of the images on the screen. Certainly, all the images from all the methods in comparison should maintain the same resolution and size. If the number of methods that are compared allows, the images should be positioned next to each other, otherwise, regardless of the number of methods, the order of the images should be randomized. It has been shown that including the content and style images as part of the questions has an impact on the results \cite{ioannou2022depth}, thus, depending on the question the participants are asked, an appropriate choice should be made regarding the inclusion of the images used to infer the stylizations. For example, if the question considers content preservation or style resemblance, the content and style images should be revealed; if the question is in the form of ``choose the favorite stylization" or ``select the image with the overall highest quality", the authors should consider carefully if it is appropriate to reveal the content and style images.

\subsubsection{Reporting Results}

Another consideration that lacks attention in the current literature is the concept of statistical significance in the presentation and reporting of the results. The results of user studies are mostly reported as aggregate preference rates for each method or the total number of votes accumulated for each method during the comparison, or the proportion of times the method in question was selected in comparison to each of the other methods. Yet, as is essential in literature in the area of psychology, user studies should utilize statistical tests that can better interpret the collected data. This can account for the inherent uncertainty in the results, considering that various factors such as participants' reliability, the number of participants, the number of questions asked per participant, and the experimental setting can influence the reproducibility of the findings.

The concept of statistical significance is used in psychological research to help determine whether the differences or relationships we observe in data are statistically meaningful or if they could have occurred by chance \cite{lykken1968statistical}. There is a range of statistical tests that can be carried out, most of them being suitable for paired data. A suggested approach for detecting statistical significance in the results of a user study and determining the best approach amongst the compared methods is depicted in Table~\ref{tab:reporting_results}.

The concept of statistical power can also be useful. Statistical power is the probability that the test correctly rejects the null hypothesis \cite{ellis2010essential}; reporting this by complementing the statistical tests can allow for a better understanding of the significance of the results. It is also a practical way to estimate the minimum sample size required for an experiment.

\begin{table}[htb]
\centering
    \caption{Suggested approach for detecting statistical significance in user studies' results.}
    \label{tab:reporting_results}

    \begin{tabular}{p{0.155\linewidth}  p{0.75\linewidth}}
        \hline 
        
      \multirow{4}{*}{\textbf{Setup}}  &   Collecting data from multiple participants results in an $N \times N$ contingency table where each cell represents the number of times a particular method was chosen to produce the favorite image over the other $N-1$ methods.  \\ \hline
        
        \multirow{7}{*}{\textbf{Hypothesis}} &  The null hypothesis ($H_{0}$) would be that there is no significant difference between the methods in terms of being chosen to produce the most preferred synthesized stylizations. The alternative hypothesis ($H_{1}$) would be that there is a significant difference between the performance of the approaches in comparison. A thorough explanation of hypothesis testing is given in \cite{duca2022}.  \\ \hline
        
        \multirow{17}{=}{\textbf{Statistical Tests}} 

        & \underline{\textit{t-test:}} A commonly used statistical test that compares the means of the two samples. One approach is to perform a series of pairwise t-tests to determine if there are significant differences between each pair of methods. Essentially, this would mean comparing the performance of the suggested NST technique with the other state-of-the-art methods. \\ 
        & \underline{\textit{Rank aggregation:}} The overall ranking of each method can be calculated based on the number of times it was chosen as the favorite across all questions. Statistical methods such as the Friedman test \cite{friedman1937use} or the Wilcoxon signed-rank test \cite{wilcoxon1992individual}  can be employed to determine if there are significant differences in the rankings amongst methods. \\ 
        &  \underline{\textit{Effect size:}} Effect size measures such as Cohen's d \cite{cohen2013statistical} or Cliff's delta \cite{cliff1993dominance} can be used to quantify the magnitude of the differences between methods, providing extra information about the practical significance of the findings. \\ \hline

        \multirow{13}{=}{\textbf{Documenting the results}} & 
        
         \underline{\textit{Statistical test results:}} Reporting the p-values obtained from the t-tests for each pair of methods could enhance the reliability of the results. Any significant differences found can also be highlighted. \\

        & \underline{\textit{Effect sizes:}} Effect size measures can be included to provide a more comprehensive understanding of the differences between methods. \\

        & \underline{\textit{Limitations:}} An essential part of the presentation of the results is the discussion of any limitations of the study, such as sample size, potential biases, or specific characteristics of the images or methods used. \\

        & \underline{\textit{Visual aids:}} Using graphs or tables to present the results will allow readers to understand the findings more easily. \\ \hline      
                
    \end{tabular}
\end{table}

\subsection{Quantitative Evaluation Metrics}
\label{sec:analysis-quantitative}

There is no consensus on the most suitable quantitative metrics to use in NST studies. Nonetheless, as discussed in Section~\ref{sec:current-evaluation}, we can divide them into four categories depending on the aspect of the NST process they assess: perceptual metrics, stylization performance, video metrics, and efficiency. Speed and memory comparisons make up the efficiency category. However, there is a range of different metrics that compose each of the other categories. Different metrics examine specific aspects, and in some cases, even the same aspects, but employ distinct methodologies to assess them. 

To suggest only a selection of those metrics as the benchmark evaluation methodology is a challenging task. However, it is a necessary task as it would be extremely inefficient to quantify the performance of an NST system by employing all the metrics depicted in Figure~\ref{fig:evaluation-taxonomy}. Depending on the aim of a method, a few of the metrics can be discarded. For example, for a study proposing an algorithm for improving the stylization effect or attempting to produce an aesthetic effect \cite{hu2020aesthetic}, computing depth map or edge map differences may not be appropriate. Other metrics such as Content and Style error may also be discarded if seen from a more critical lens. As these are defined and utilized in the training loop, they are not providing anything useful -- their use points to a circular definition since the evaluation of the content preservation or style performance qualities of the stylized results relies on the same means that were employed to create them. Future work will be required to necessitate a more in-depth analysis of the employed computational metrics, scrutinising their utility and performance, with the ultimate aim of establishing a universally adopted benchmark array of metrics, consisting of a singular metric per aspect of the evaluated NST performance. Such an effort would contribute to standardising the evaluation process and promoting consistent and meaningful comparisons among different NST methods.

As this could be a challenging task, statistical analysis, as described in the previous section regarding User Studies, can aid in the validity and reliability of the reported results. For any computational metric that is employed, instead of the computation of an average of the performance of each method on a test dataset, again, pairwise t-tests can be utilized to dictate the statistical significance, and, as a consequence, the reliability of the results. This would be useful, particularly if a benchmark dataset is not universally adopted --  the statistical tests can provide a useful approximation of the repeatability of the results for different data.

Recently, several methods have considered evaluation in NST, proposing new quantitative metrics. The approach of Chen~\etal \cite{chen2023learning} is capable of assessing how well a stylized image resembles an authentic artwork. The system suggested by Wright and Ommer \cite{wright2022artfid} combines both content preservation and style matching evaluation in a single metric, namely \textit{ArtFID}. Also, Chen~\etal \cite{chen2023collaborative} developed a network (\textit{CLSAP-Net}) that uses collaborative learning which is composed of a content preservation estimation network (CPE-Net), a style resemblance estimation network (SRE-Net), and an overall vision target network (OVT-Net) attempting to provide a robust metric that effectively assesses all aspects of stylization.

These metrics have exhibited a significant correlation with human judgment, suggesting that the reliance on user studies for subjective judgments can be alleviated. These efforts hold great promise in the field of NST, offering a pathway to establish a standardized evaluation procedure that reduces the reliance on costly and potentially flawed user surveys.

Although the definition of NST does not explicitly consider the aesthetics of the results, arguably, this is a concept encapsulated in the artistic nature of the stylizations. The goal of NST is to create artistic imagery, and often the recruited participants in the setting of a user study are asked to choose their favorite artistic synthesis. Despite the content and style considerations, the selection of a ``favorite" image is also based on what the viewer considers to be aesthetically pleasing. The aesthetics of the NST results is something that requires further investigation in order to adequately capture the full spectrum of the human subjective experience and ultimately replace the widely used user studies. The recently developed computational aesthetic evaluation field \cite{murray2012ava,mohammad2018wikiart,achlioptas2021artemis,talebi2018nima,yi2023towards} encompasses conventional (based on hand-crafted features) and learning-based (based on Deep Learning) approaches that attempt to derive aesthetic judgments on images. As the aims of the computational aesthetic evaluation field strongly correlate with the aims of NST, resorting to such approaches could be catalytic in converging to a more robust quantitative evaluation that considers the aesthetics of the results.

\subsection{Summary of Evaluation Issues and Recommendations}

An exhaustive analysis of the state-of-the-art methods has led to a collection of recommendations that could alleviate the present evaluation issues and serve as a basis for the development of a standardized evaluation protocol. Table~\ref{tbl:summary} provides a condensed summary of the prevailing evaluation challenges and our recommendations to address them.


\begin{table*}[htb]
\setlength\extrarowheight{1pt}

 \centering
    \caption{Summary of Issues in Evaluation and Recommendations}\label{tbl:summary}

     \resizebox{1\textwidth}{!}{%
        
     \begin{tabular}{ p{4.5em} p{55em} }

    \toprule
        \textbf{\begin{tabular}[c]{@{}l@{}}Evaluation\\Aspect\end{tabular}} & \textbf{Issues [I] \& Recommendations [R]} \\ 
    
    \midrule 

       
       Datasets & 
       [I]: No universally agreed benchmark evaluation dataset exists. Different data is used for each type of evaluation.  \newline
       [R]: Development of benchmark evaluation datasets for the different aspects of evaluation: for Qualitative Evaluation, a small set of content images/video frames and style images that is universally shared and presented for each NST study; for User Studies, a set of content and style images/video frames that capture a wide variety of characteristics and suffice for the development of adequate questions; for Quantitative Evaluation, a large-scale test dataset to provide more accurate and reproducible results. \\ \hline
       
        Qualitative Evaluation & 
        [I]: Inconsistencies in the data presented and commented on. Lack of common practices in deriving subjective judgements.  \newline
        [R]: A small dataset to be commonly shared among the methods. The subjective judgments to be clearly focused on addressing essential aspects, such as content preservation, stylization performance, observed artifacts, and overall aesthetic quality. \\ \hline

        Human Evaluation Studies & 
        [I]: Dissimilarities in the design and formulation of user studies, and consequently in the presentation of the results. Lack of consensus regarding the data utilized, the number and background of participants recruited, the quantity and presentation of the questions, and the reporting of the results. \newline
        [R]: Development of a standardized practice for conducting user surveys. The different aspects of human evaluation studies should be considered as they can significantly influence the results and their interpretation. Essential aspects include the data that is utilized, the recruitment of participants, the number of methods in a comparison, the presentation of the questions and their formulation, the reporting of the results and potential statistical analysis.
        \\ \hline

        Quantitative Evaluation Metrics & 
        [I]: Different methods employ different metrics. Absence of a standardized set of quantitative evaluation metrics. \newline 
        [R]: Achieve a consensus on a handful of metrics that can adequately and effectively evaluate the NST results. Resort to the field of Computational Aesthetics to allow for metrics that can quantify not only content preservation and stylization performance but also the aesthetic quality of the results.  \\ 
      
    \bottomrule
    \end{tabular}%
    }
\end{table*}

\section{Conclusion}

The success and widespread adoption of deep learning and computer vision techniques has undoubtedly influenced and shaped the trajectory of NST research. Along with the unparalleled success of diffusion models and text-to-image generative approaches, the field of image and video neural style transfer has continued to progress and produce remarkable ideas and implementations. Yet, the considerable progress and maturation of the field over the years has not led to the establishment of a robust and reliable evaluation protocol.

This review paper has examined the different evaluation techniques in image and video NST literature.
Qualitative evaluations offer useful insights but suffer from subjectivity and reproducibility issues. Human evaluation studies, while subjective, gather quantitative data through user surveys with inconsistent methodologies across NST methods. Quantitative evaluation includes multiple computational metrics that can reliably assess the various facets of the performance of NST algorithms. Despite being reproducible and repeatable, the quantitative computational metrics are not utilized identically by each NST method. Different data and different metrics are selectively employed by each approach, suggesting a requirement for a universally agreed benchmark protocol.


Another point of contention relates to the use of data in the evaluation process. No benchmark evaluation dataset exists and the nature of the diverse evaluation approaches further exacerbates the complexity of this issue. In this paper, we have underscored the challenges arising from the utilization of disparate and often undisclosed or undefined data sources in the NST evaluation methods and emphasized the significance of establishing benchmark test datasets. Such datasets would not only facilitate fair comparisons among different methods but also enhance the transparency and reproducibility of the reported results.

The rapid advancements in NST techniques and the proliferation of diverse approaches have created a diverse landscape of evaluation methods and metrics. However, there is a lack of consensus on the most appropriate and effective evaluation criteria, leading to inconsistencies and limitations in the evaluation process. This inherent complexity highlights the need for a comprehensive and standardized evaluation protocol that can effectively assess and compare the performance of different NST methods. Such a protocol would provide a solid foundation for advancing the field, ensuring the credibility of research findings, and facilitating meaningful comparisons between different approaches. 
The analysis in our paper provides a foundation upon which a standardized evaluation process could be designed and developed. A universal evaluation framework 
could ensure the reliability, repeatability and reproducibility required for further robust developments in the NST landscape.





 
%

\bibliographystyle{IEEEtran}
\bibliography{bib/IEEEabrv,bib/references}

\begin{thebibliography}{100}
\providecommand{\url}[1]{#1}
\csname url@samestyle\endcsname
\providecommand{\newblock}{\relax}
\providecommand{\bibinfo}[2]{#2}
\providecommand{\BIBentrySTDinterwordspacing}{\spaceskip=0pt\relax}
\providecommand{\BIBentryALTinterwordstretchfactor}{4}
\providecommand{\BIBentryALTinterwordspacing}{\spaceskip=\fontdimen2\font plus
\BIBentryALTinterwordstretchfactor\fontdimen3\font minus \fontdimen4\font\relax}
\providecommand{\BIBforeignlanguage}[2]{{%
\expandafter\ifx\csname l@#1\endcsname\relax
\typeout{** WARNING: IEEEtran.bst: No hyphenation pattern has been}%
\typeout{** loaded for the language `#1'. Using the pattern for}%
\typeout{** the default language instead.}%
\else
\language=\csname l@#1\endcsname
\fi
#2}}
\providecommand{\BIBdecl}{\relax}
\BIBdecl

\bibitem{gatys2015neural}
L.~A. Gatys, A.~S. Ecker, and M.~Bethge, ``A neural algorithm of artistic style,'' \emph{arXiv preprint arXiv:1508.06576}, 2015.

\bibitem{jing2019neural}
Y.~Jing, Y.~Yang, Z.~Feng, J.~Ye, Y.~Yu, and M.~Song, ``Neural style transfer: A review,'' \emph{IEEE transactions on visualization and computer graphics}, 2019.

\bibitem{liu2021learning}
X.-C. Liu, Y.-L. Yang, and P.~Hall, ``Learning to warp for style transfer,'' in \emph{Proceedings of the IEEE/CVF Conference on Computer Vision and Pattern Recognition}, 2021, pp. 3702--3711.

\bibitem{sanakoyeu2018style}
A.~Sanakoyeu, D.~Kotovenko, S.~Lang, and B.~Ommer, ``A style-aware content loss for real-time hd style transfer,'' in \emph{Proceedings of the European Conference on Computer Vision (ECCV)}, 2018, pp. 698--714.

\bibitem{singh2021neural}
A.~Singh, V.~Jaiswal, G.~Joshi, A.~Sanjeeve, S.~Gite, and K.~Kotecha, ``Neural style transfer: A critical review,'' \emph{IEEE Access}, vol.~9, pp. 131\,583--131\,613, 2021.

\bibitem{radford2021learning}
A.~Radford, J.~W. Kim, C.~Hallacy, A.~Ramesh, G.~Goh, S.~Agarwal, G.~Sastry, A.~Askell, P.~Mishkin, J.~Clark \emph{et~al.}, ``Learning transferable visual models from natural language supervision,'' in \emph{International conference on machine learning}.\hskip 1em plus 0.5em minus 0.4em\relax PMLR, 2021, pp. 8748--8763.

\bibitem{gatys2017controlling}
L.~A. Gatys, A.~S. Ecker, M.~Bethge, A.~Hertzmann, and E.~Shechtman, ``Controlling perceptual factors in neural style transfer,'' in \emph{Proceedings of the IEEE Conference on Computer Vision and Pattern Recognition}, 2017, pp. 3985--3993.

\bibitem{Li2017demistifying}
Y.~Li, N.~Wang, J.~Liu, and X.~Hou, ``Demystifying neural style transfer,'' in \emph{Proceedings of the 26th International Joint Conference on Artificial Intelligence}, ser. IJCAI'17.\hskip 1em plus 0.5em minus 0.4em\relax AAAI Press, 2017, p. 2230–2236.

\bibitem{Risser2017stable}
\BIBentryALTinterwordspacing
E.~Risser, P.~Wilmot, and C.~Barnes, ``Stable and controllable neural texture synthesis and style transfer using histogram losses,'' \emph{arXiv preprint arXiv:1701.08893}, 1 2017. [Online]. Available: \url{https://arxiv.org/abs/1701.08893v2}
\BIBentrySTDinterwordspacing

\bibitem{Li2017laplacian}
\BIBentryALTinterwordspacing
S.~Li, X.~Xu, L.~Nie, and T.-S. Chua, ``Laplacian-steered neural style transfer,'' in \emph{Proceedings of the 25th ACM international conference on Multimedia}, 2017, pp. 1716--1724. [Online]. Available: \url{https://doi.org/10.1145/3123266.3123425}
\BIBentrySTDinterwordspacing

\bibitem{li2016combining}
C.~Li and M.~Wand, ``Combining markov random fields and convolutional neural networks for image synthesis,'' in \emph{Proceedings of the IEEE Conference on Computer Vision and Pattern Recognition (CVPR)}, June 2016.

\bibitem{Luan2017deep}
\BIBentryALTinterwordspacing
F.~Luan, S.~Paris, E.~Shechtman, and K.~Bala, ``Deep photo style transfer,'' in \emph{Proceedings of the IEEE conference on computer vision and pattern recognition}, 3 2017, pp. 4990--4998. [Online]. Available: \url{http://arxiv.org/abs/1703.07511}
\BIBentrySTDinterwordspacing

\bibitem{mechrez2017photorealistic}
R.~Mechrez, E.~Shechtman, and L.~Zelnik-Manor, ``Photorealistic style transfer with screened poisson equation,'' \emph{arXiv preprint arXiv:1709.09828}, 2017.

\bibitem{penhouet2019automated}
\BIBentryALTinterwordspacing
S.~Penhouët and P.~Sanzenbacher, ``Automated deep photo style transfer,'' \emph{arXiv preprint arXiv:1901.03915}, 1 2019. [Online]. Available: \url{http://arxiv.org/abs/1901.03915}
\BIBentrySTDinterwordspacing

\bibitem{gatys2016image}
L.~A. Gatys, A.~S. Ecker, and M.~Bethge, ``Image style transfer using convolutional neural networks,'' in \emph{Proceedings of the IEEE conference on computer vision and pattern recognition}, 2016, pp. 2414--2423.

\bibitem{johnson2016perceptual}
J.~Johnson, A.~Alahi, and L.~Fei-Fei, ``Perceptual losses for real-time style transfer and super-resolution,'' in \emph{European conference on computer vision}.\hskip 1em plus 0.5em minus 0.4em\relax Springer, 2016, pp. 694--711.

\bibitem{ulyanov2016texture}
D.~Ulyanov, V.~Lebedev, A.~Vedaldi, and V.~S. Lempitsky, ``Texture networks: Feed-forward synthesis of textures and stylized images.'' in \emph{ICML}, vol.~1, 2016, p.~4.

\bibitem{ulyanov2017improved}
D.~Ulyanov, A.~Vedaldi, and V.~Lempitsky, ``Improved texture networks: Maximizing quality and diversity in feed-forward stylization and texture synthesis,'' in \emph{Proceedings of the IEEE Conference on Computer Vision and Pattern Recognition}, 2017, pp. 6924--6932.

\bibitem{ulyanov2016instance}
D.~Ulyanov, A.~Vedaldi, and V.~S. Lempitsky, ``Instance normalization: The missing ingredient for fast tylization,'' 2017.

\bibitem{li2016precomputed}
C.~Li and M.~Wand, ``Precomputed real-time texture synthesis with markovian generative adversarial networks,'' in \emph{Computer Vision--ECCV 2016: 14th European Conference, Amsterdam, The Netherlands, October 11-14, 2016, Proceedings, Part III 14}.\hskip 1em plus 0.5em minus 0.4em\relax Springer, 2016, pp. 702--716.

\bibitem{liu2017depth}
X.-C. Liu, M.-M. Cheng, Y.-K. Lai, and P.~L. Rosin, ``Depth-aware neural style transfer,'' in \emph{Proceedings of the Symposium on Non-Photorealistic Animation and Rendering}, 2017, pp. 1--10.

\bibitem{cheng2019structure}
M.-M. Cheng, X.-C. Liu, J.~Wang, S.-P. Lu, Y.-K. Lai, and P.~L. Rosin, ``Structure-preserving neural style transfer,'' \emph{IEEE Transactions on Image Processing}, vol.~29, pp. 909--920, 2019.

\bibitem{ioannou2022depth}
E.~Ioannou and S.~Maddock, ``Depth-aware neural style transfer using instance normalization,'' in \emph{Computer Graphics \& Visual Computing (CGVC) 2022}.\hskip 1em plus 0.5em minus 0.4em\relax Eurographics Digital Library, 2022.

\bibitem{kotovenko2019content}
D.~Kotovenko, A.~Sanakoyeu, S.~Lang, and B.~Ommer, ``Content and style disentanglement for artistic style transfer,'' in \emph{Proceedings of the IEEE/CVF international conference on computer vision}, 2019, pp. 4422--4431.

\bibitem{dumoulin2017learned}
\BIBentryALTinterwordspacing
V.~Dumoulin, J.~Shlens, and M.~Kudlur, ``A learned representation for artistic style,'' in \emph{International Conference on Learning Representations}, 2017. [Online]. Available: \url{https://openreview.net/forum?id=BJO-BuT1g}
\BIBentrySTDinterwordspacing

\bibitem{chen2017stylebank}
D.~Chen, L.~Yuan, J.~Liao, N.~Yu, and G.~Hua, ``Stylebank: An explicit representation for neural image style transfer,'' in \emph{Proceedings of the IEEE conference on computer vision and pattern recognition}, 2017, pp. 1897--1906.

\bibitem{zhang2018multi}
H.~Zhang and K.~Dana, ``Multi-style generative network for real-time transfer,'' in \emph{Proceedings of the European Conference on Computer Vision (ECCV) Workshops}, 2018, pp. 0--0.

\bibitem{li2017diversified}
Y.~Li, C.~Fang, J.~Yang, Z.~Wang, X.~Lu, and M.-H. Yang, ``Diversified texture synthesis with feed-forward networks,'' in \emph{Proceedings of the IEEE conference on computer vision and pattern recognition}, 2017, pp. 3920--3928.

\bibitem{chen2016fast}
T.~Q. Chen and M.~Schmidt, ``Fast patch-based style transfer of arbitrary style,'' \emph{arXiv preprint arXiv:1612.04337}, 2016.

\bibitem{huang2017arbitrary}
X.~Huang and S.~Belongie, ``Arbitrary style transfer in real-time with adaptive instance normalization,'' in \emph{Proceedings of the IEEE International Conference on Computer Vision}, 2017, pp. 1501--1510.

\bibitem{ghiasi2017exploring}
G.~Ghiasi, H.~Lee, M.~Kudlur, V.~Dumoulin, and J.~Shlens, ``Exploring the structure of a real-time, arbitrary neural artistic stylization network,'' \emph{arXiv preprint arXiv:1705.06830}, 2017.

\bibitem{li2017universal}
Y.~Li, C.~Fang, J.~Yang, Z.~Wang, X.~Lu, and M.-H. Yang, ``Universal style transfer via feature transforms,'' \emph{Advances in neural information processing systems}, vol.~30, 2017.

\bibitem{xu2018learning}
Z.~Xu, M.~Wilber, C.~Fang, A.~Hertzmann, and H.~Jin, ``Learning from multi-domain artistic images for arbitrary style transfer,'' \emph{arXiv preprint arXiv:1805.09987}, 2018.

\bibitem{huo2021manifold}
J.~Huo, S.~Jin, W.~Li, J.~Wu, Y.-K. Lai, Y.~Shi, and Y.~Gao, ``Manifold alignment for semantically aligned style transfer,'' in \emph{Proceedings of the IEEE/CVF International Conference on Computer Vision}, 2021, pp. 14\,861--14\,869.

\bibitem{park2019arbitrary}
D.~Y. Park and K.~H. Lee, ``Arbitrary style transfer with style-attentional networks,'' in \emph{proceedings of the IEEE/CVF conference on computer vision and pattern recognition}, 2019, pp. 5880--5888.

\bibitem{svoboda2020two}
J.~Svoboda, A.~Anoosheh, C.~Osendorfer, and J.~Masci, ``Two-stage peer-regularized feature recombination for arbitrary image style transfer,'' in \emph{Proceedings of the IEEE/CVF Conference on Computer Vision and Pattern Recognition}, 2020, pp. 13\,816--13\,825.

\bibitem{an2020ultrafast}
\BIBentryALTinterwordspacing
J.~An, H.~Xiong, J.~Huan, and J.~Luo, ``Ultrafast photorealistic style transfer via neural architecture search,'' \emph{Proceedings of the AAAI Conference on Artificial Intelligence}, vol.~34, no.~07, pp. 10\,443--10\,450, Apr. 2020. [Online]. Available: \url{https://ojs.aaai.org/index.php/AAAI/article/view/6614}
\BIBentrySTDinterwordspacing

\bibitem{hu2020aesthetic}
Z.~Hu, J.~Jia, B.~Liu, Y.~Bu, and J.~Fu, ``Aesthetic-aware image style transfer,'' in \emph{Proceedings of the 28th ACM International Conference on Multimedia}, 2020, pp. 3320--3329.

\bibitem{hong2023aespa}
K.~Hong, S.~Jeon, J.~Lee, N.~Ahn, K.~Kim, P.~Lee, D.~Kim, Y.~Uh, and H.~Byun, ``{AesPA-Net}: Aesthetic pattern-aware style transfer networks,'' in \emph{Proceedings of the IEEE/CVF International Conference on Computer Vision}, 2023, pp. 22\,758--22\,767.

\bibitem{shen2018neural}
F.~Shen, S.~Yan, and G.~Zeng, ``Neural style transfer via meta networks,'' in \emph{Proceedings of the IEEE Conference on Computer Vision and Pattern Recognition}, 2018, pp. 8061--8069.

\bibitem{liu2021adaattn}
S.~Liu, T.~Lin, D.~He, F.~Li, M.~Wang, X.~Li, Z.~Sun, Q.~Li, and E.~Ding, ``Adaattn: Revisit attention mechanism in arbitrary neural style transfer,'' in \emph{Proceedings of the IEEE/CVF international conference on computer vision}, 2021, pp. 6649--6658.

\bibitem{deng2022stytr2}
Y.~Deng, F.~Tang, W.~Dong, C.~Ma, X.~Pan, L.~Wang, and C.~Xu, ``Stytr2: Image style transfer with transformers,'' in \emph{Proceedings of the IEEE/CVF Conference on Computer Vision and Pattern Recognition}, 2022, pp. 11\,326--11\,336.

\bibitem{luo2022consistent}
X.~Luo, Z.~Han, L.~Yang, and L.~Zhang, ``Consistent style transfer,'' \emph{arXiv preprint arXiv:2201.02233}, 2022.

\bibitem{zhu2023all}
M.~Zhu, X.~He, N.~Wang, X.~Wang, and X.~Gao, ``All-to-key attention for arbitrary style transfer,'' in \emph{Proceedings of the IEEE/CVF International Conference on Computer Vision}, 2023, pp. 23\,109--23\,119.

\bibitem{zhang2023unified}
Y.~Zhang, F.~Tang, W.~Dong, H.~Huang, C.~Ma, T.-Y. Lee, and C.~Xu, ``A unified arbitrary style transfer framework via adaptive contrastive learning,'' \emph{ACM Transactions on Graphics}, 2023.

\bibitem{huang2023quantart}
S.~Huang, J.~An, D.~Wei, J.~Luo, and H.~Pfister, ``Quantart: Quantizing image style transfer towards high visual fidelity,'' in \emph{Proceedings of the IEEE/CVF Conference on Computer Vision and Pattern Recognition}, 2023, pp. 5947--5956.

\bibitem{ma2023rast}
Y.~Ma, C.~Zhao, X.~Li, and A.~Basu, ``{RAST}: Restorable arbitrary style transfer via multi-restoration,'' in \emph{Proceedings of the IEEE/CVF Winter Conference on Applications of Computer Vision}, 2023, pp. 331--340.

\bibitem{li2018closed}
Y.~Li, M.-Y. Liu, X.~Li, M.-H. Yang, and J.~Kautz, ``A closed-form solution to photorealistic image stylization,'' in \emph{Proceedings of the European conference on computer vision (ECCV)}, 2018, pp. 453--468.

\bibitem{yoo2019photorealistic}
J.~Yoo, Y.~Uh, S.~Chun, B.~Kang, and J.-W. Ha, ``Photorealistic style transfer via wavelet transforms,'' in \emph{Proceedings of the IEEE/CVF International Conference on Computer Vision}, 2019, pp. 9036--9045.

\bibitem{ke2023neural}
Z.~Ke, Y.~Liu, L.~Zhu, N.~Zhao, and R.~W. Lau, ``Neural preset for color style transfer,'' in \emph{Proceedings of the IEEE/CVF Conference on Computer Vision and Pattern Recognition}, 2023, pp. 14\,173--14\,182.

\bibitem{gu2018arbitrary}
S.~Gu, C.~Chen, J.~Liao, and L.~Yuan, ``Arbitrary style transfer with deep feature reshuffle,'' in \emph{Proceedings of the IEEE Conference on Computer Vision and Pattern Recognition}, 2018, pp. 8222--8231.

\bibitem{li2019learning}
X.~Li, S.~Liu, J.~Kautz, and M.-H. Yang, ``Learning linear transformations for fast image and video style transfer,'' in \emph{Proceedings of the IEEE/CVF Conference on Computer Vision and Pattern Recognition}, 2019, pp. 3809--3817.

\bibitem{an2021artflow}
J.~An, S.~Huang, Y.~Song, D.~Dou, W.~Liu, and J.~Luo, ``Artflow: Unbiased image style transfer via reversible neural flows,'' in \emph{Proceedings of the IEEE/CVF Conference on Computer Vision and Pattern Recognition}, 2021, pp. 862--871.

\bibitem{liu2021structure}
S.~Liu and T.~Zhu, ``Structure-guided arbitrary style transfer for artistic image and video,'' \emph{IEEE Transactions on Multimedia}, 2021.

\bibitem{chen2021artistic}
H.~Chen, Z.~Wang, H.~Zhang, Z.~Zuo, A.~Li, W.~Xing, D.~Lu \emph{et~al.}, ``Artistic style transfer with internal-external learning and contrastive learning,'' \emph{Advances in Neural Information Processing Systems}, vol.~34, pp. 26\,561--26\,573, 2021.

\bibitem{wang2022aesust}
Z.~Wang, Z.~Zhang, L.~Zhao, Z.~Zuo, A.~Li, W.~Xing, and D.~Lu, ``{AesUST}: towards aesthetic-enhanced universal style transfer,'' in \emph{Proceedings of the 30th ACM International Conference on Multimedia}, 2022, pp. 1095--1106.

\bibitem{ruta2023neat}
D.~Ruta, A.~Gilbert, J.~Collomosse, E.~Shechtman, and N.~Kolkin, ``Neat: Neural artistic tracing for beautiful style transfer,'' \emph{arXiv preprint arXiv:2304.05139}, 2023.

\bibitem{xu2023learning}
W.~Xu, C.~Long, and Y.~Nie, ``Learning dynamic style kernels for artistic style transfer,'' in \emph{Proceedings of the IEEE/CVF Conference on Computer Vision and Pattern Recognition}, 2023, pp. 10\,083--10\,092.

\bibitem{tang2023master}
H.~Tang, S.~Liu, T.~Lin, S.~Huang, F.~Li, D.~He, and X.~Wang, ``Master: Meta style transformer for controllable zero-shot and few-shot artistic style transfer,'' in \emph{Proceedings of the IEEE/CVF Conference on Computer Vision and Pattern Recognition}, 2023, pp. 18\,329--18\,338.

\bibitem{gu2023two}
B.~Gu, H.~Fan, and L.~Zhang, ``Two birds, one stone: A unified framework for joint learning of image and video style transfers,'' \emph{arXiv preprint arXiv:2304.11335}, 2023.

\bibitem{li2023frequency}
D.~Li, H.~Luo, P.~Wang, Z.~Wang, S.~Liu, and F.~Wang, ``Frequency domain disentanglement for arbitrary neural style transfer,'' in \emph{Proceedings of the AAAI Conference on Artificial Intelligence}, vol.~37, no.~1, 2023, pp. 1287--1295.

\bibitem{ruder2016artistic}
M.~Ruder, A.~Dosovitskiy, and T.~Brox, ``Artistic style transfer for videos,'' in \emph{Pattern Recognition}, B.~Rosenhahn and B.~Andres, Eds.\hskip 1em plus 0.5em minus 0.4em\relax Cham: Springer International Publishing, 2016, pp. 26--36.

\bibitem{ruder2018artistic}
M.~Ruder, A.~Dosovitskiy, and T.~Brox", ``Artistic style transfer for videos and spherical images,'' \emph{International Journal of Computer Vision}, vol. 126, no.~11, pp. 1199--1219, 2018.

\bibitem{huang2017real}
H.~Huang, H.~Wang, W.~Luo, L.~Ma, W.~Jiang, X.~Zhu, Z.~Li, and W.~Liu, ``Real-time neural style transfer for videos,'' in \emph{Proceedings of the IEEE Conference on Computer Vision and Pattern Recognition}, 2017, pp. 783--791.

\bibitem{gao2018reconet}
C.~Gao, D.~Gu, F.~Zhang, and Y.~Yu, ``{ReCoNet}: Real-time coherent video style transfer network,'' 2018.

\bibitem{ioannou2023depth}
E.~Ioannou and S.~Maddock, ``Depth-aware neural style transfer for videos,'' \emph{Computers}, vol.~12, no.~4, p.~69, 2023.

\bibitem{gao2020fast}
W.~Gao, Y.~Li, Y.~Yin, and M.-H. Yang, ``Fast video multi-style transfer,'' in \emph{Proceedings of the IEEE/CVF winter conference on applications of computer vision}, 2020, pp. 3222--3230.

\bibitem{wang2020consistent}
W.~Wang, S.~Yang, J.~Xu, and J.~Liu, ``Consistent video style transfer via relaxation and regularization,'' \emph{IEEE Transactions on Image Processing}, vol.~29, pp. 9125--9139, 2020.

\bibitem{deng2021arbitrary}
\BIBentryALTinterwordspacing
Y.~Deng, F.~Tang, W.~Dong, H.~Huang, C.~Ma, and C.~Xu, ``Arbitrary video style transfer via multi-channel correlation,'' \emph{Proceedings of the AAAI Conference on Artificial Intelligence}, vol.~35, no.~2, pp. 1210--1217, May 2021. [Online]. Available: \url{https://ojs.aaai.org/index.php/AAAI/article/view/16208}
\BIBentrySTDinterwordspacing

\bibitem{lu2022universal}
H.~Lu and Z.~Wang, ``Universal video style transfer via crystallization, separation, and blending,'' in \emph{Proc. Int. Joint Conf. on Artif. Intell.(IJCAI)}, vol.~36, 2022, pp. 4957--4965.

\bibitem{wu2022ccpl}
Z.~Wu, Z.~Zhu, J.~Du, and X.~Bai, ``{CCPL}: Contrastive coherence preserving loss for versatile style transfer,'' in \emph{Computer Vision--ECCV 2022: 17th European Conference, Tel Aviv, Israel, October 23--27, 2022, Proceedings, Part XVI}.\hskip 1em plus 0.5em minus 0.4em\relax Springer, 2022, pp. 189--206.

\bibitem{xia2021real}
X.~Xia, T.~Xue, W.-s. Lai, Z.~Sun, A.~Chang, B.~Kulis, and J.~Chen, ``Real-time localized photorealistic video style transfer,'' in \emph{Proceedings of the IEEE/CVF Winter Conference on Applications of Computer Vision}, 2021, pp. 1089--1098.

\bibitem{kurzman2019classbased}
L.~Kurzman, D.~Vazquez, and I.~Laradji, ``Class-based styling: Real-time localized style transfer with semantic segmentation,'' 2019.

\bibitem{Jamriska19-SIG}
O.~Jamri\v{s}ka, \v{S}\'{a}rka Sochorov\'{a}, O.~Texler, M.~Luk\'{a}\v{c}, J.~Fi\v{s}er, J.~Lu, E.~Shechtman, and D.~S\'{y}kora, ``Stylizing video by example,'' \emph{ACM Transactions on Graphics}, vol.~38, no.~4, 2019.

\bibitem{chen2023learning}
J.~Chen, J.~An, H.~Lyu, and J.~Luo, ``Learning to evaluate the artness of ai-generated images,'' \emph{arXiv preprint arXiv:2305.04923}, 2023.

\bibitem{lin2014microsoft}
T.-Y. Lin, M.~Maire, S.~Belongie, J.~Hays, P.~Perona, D.~Ramanan, P.~Doll{\'a}r, and C.~L. Zitnick, ``Microsoft coco: Common objects in context,'' in \emph{European conference on computer vision}.\hskip 1em plus 0.5em minus 0.4em\relax Springer, 2014, pp. 740--755.

\bibitem{mohammad2018wikiart}
S.~Mohammad and S.~Kiritchenko, ``Wikiart emotions: An annotated dataset of emotions evoked by art,'' in \emph{Proceedings of the eleventh international conference on language resources and evaluation (LREC 2018)}, 2018.

\bibitem{ostagram}
\BIBentryALTinterwordspacing
{Ostagram}. [Online]. Available: \url{https://www.ostagram.me/}
\BIBentrySTDinterwordspacing

\bibitem{karayev2014recognising}
S.~Karayev, M.~Trentacoste, H.~Han, A.~Agarwala, T.~Darrell, A.~Hertzmann, and H.~Winnemoeller, ``Recognizing image style,'' in \emph{Proceedings of the British Machine Vision Conference}.\hskip 1em plus 0.5em minus 0.4em\relax BMVA Press, 2014.

\bibitem{wilber2017bam}
M.~J. Wilber, C.~Fang, H.~Jin, A.~Hertzmann, J.~Collomosse, and S.~Belongie, ``Bam! the behance artistic media dataset for recognition beyond photography,'' in \emph{Proceedings of the IEEE international conference on computer vision}, 2017, pp. 1202--1211.

\bibitem{mould2017developing}
D.~Mould and P.~L. Rosin, ``Developing and applying a benchmark for evaluating image stylization,'' \emph{Computers \& Graphics}, vol.~67, pp. 58--76, 2017.

\bibitem{Butler2012sintel}
D.~J. Butler, J.~Wulff, G.~B. Stanley, and M.~J. Black, ``A naturalistic open source movie for optical flow evaluation,'' in \emph{European Conf. on Computer Vision (ECCV)}, ser. Part IV, LNCS 7577, {A. Fitzgibbon et al. (Eds.)}, Ed.\hskip 1em plus 0.5em minus 0.4em\relax Springer-Verlag, Oct. 2012, pp. 611--625.

\bibitem{videvo2019}
\BIBentryALTinterwordspacing
Videvo, 2019. [Online]. Available: \url{https://www.videvo.net/}
\BIBentrySTDinterwordspacing

\bibitem{richter2016playing}
S.~R. Richter, V.~Vineet, S.~Roth, and V.~Koltun, ``Playing for data: Ground truth from computer games,'' in \emph{Computer Vision--ECCV 2016: 14th European Conference, Amsterdam, The Netherlands, October 11-14, 2016, Proceedings, Part II 14}.\hskip 1em plus 0.5em minus 0.4em\relax Springer, 2016, pp. 102--118.

\bibitem{pont20172017}
J.~Pont-Tuset, F.~Perazzi, S.~Caelles, P.~Arbel{\'a}ez, A.~Sorkine-Hornung, and L.~Van~Gool, ``The 2017 davis challenge on video object segmentation,'' \emph{arXiv preprint arXiv:1704.00675}, 2017.

\bibitem{kohavi2009controlled}
R.~Kohavi, R.~Longbotham, D.~Sommerfield, and R.~M. Henne, ``Controlled experiments on the web: survey and practical guide,'' \emph{Data mining and knowledge discovery}, vol.~18, pp. 140--181, 2009.

\bibitem{kitov2019depth}
V.~Kitov, K.~Kozlovtsev, and M.~Mishustina, ``Depth-aware arbitrary style transfer using instance normalization,'' \emph{arXiv preprint arXiv:1906.01123}, 2019.

\bibitem{bylinskii2023towards}
Z.~Bylinskii, L.~Herman, A.~Hertzmann, S.~Hutka, Y.~Zhang \emph{et~al.}, ``Towards better user studies in computer graphics and vision,'' \emph{Foundations and Trends{\textregistered} in Computer Graphics and Vision}, vol.~15, no.~3, pp. 201--252, 2023.

\bibitem{wang2004image}
Z.~Wang, A.~C. Bovik, H.~R. Sheikh, and E.~P. Simoncelli, ``Image quality assessment: from error visibility to structural similarity,'' \emph{IEEE transactions on image processing}, vol.~13, no.~4, pp. 600--612, 2004.

\bibitem{zhang2018unreasonable}
R.~Zhang, P.~Isola, A.~A. Efros, E.~Shechtman, and O.~Wang, ``The unreasonable effectiveness of deep features as a perceptual metric,'' in \emph{Proceedings of the IEEE conference on computer vision and pattern recognition}, 2018, pp. 586--595.

\bibitem{wang2021evaluate}
Z.~Wang, L.~Zhao, H.~Chen, Z.~Zuo, A.~Li, W.~Xing, and D.~Lu, ``Evaluate and improve the quality of neural style transfer,'' \emph{Computer Vision and Image Understanding}, vol. 207, p. 103203, 2021.

\bibitem{buchner_2021}
\BIBentryALTinterwordspacing
J.~Buchner, ``Imagehash,'' 2021. [Online]. Available: \url{https://pypi.org/project/ImageHash/}
\BIBentrySTDinterwordspacing

\bibitem{chen2016single}
W.~Chen, Z.~Fu, D.~Yang, and J.~Deng, ``Single-image depth perception in the wild,'' \emph{arXiv preprint arXiv:1604.03901}, 2016.

\bibitem{Ranftl2020}
R.~Ranftl, K.~Lasinger, D.~Hafner, K.~Schindler, and V.~Koltun, ``Towards robust monocular depth estimation: Mixing datasets for zero-shot cross-dataset transfer,'' \emph{IEEE Transactions on Pattern Analysis and Machine Intelligence (TPAMI)}, 2020.

\bibitem{xie2015holistically}
S.~Xie and Z.~Tu, ``Holistically-nested edge detection,'' in \emph{Proceedings of the IEEE international conference on computer vision}, 2015, pp. 1395--1403.

\bibitem{liu2017richer}
Y.~Liu, M.-M. Cheng, X.~Hu, K.~Wang, and X.~Bai, ``Richer convolutional features for edge detection,'' in \emph{Proceedings of the IEEE conference on computer vision and pattern recognition}, 2017, pp. 3000--3009.

\bibitem{jiang2013salient}
H.~Jiang, J.~Wang, Z.~Yuan, Y.~Wu, N.~Zheng, and S.~Li, ``Salient object detection: A discriminative regional feature integration approach,'' in \emph{2013 IEEE Conference on Computer Vision and Pattern Recognition}, 2013, pp. 2083--2090.

\bibitem{ignatov2017dslr}
A.~Ignatov, N.~Kobyshev, R.~Timofte, K.~Vanhoey, and L.~Van~Gool, ``Dslr-quality photos on mobile devices with deep convolutional networks,'' in \emph{Proceedings of the IEEE international conference on computer vision}, 2017, pp. 3277--3285.

\bibitem{heusel2017gans}
M.~Heusel, H.~Ramsauer, T.~Unterthiner, B.~Nessler, and S.~Hochreiter, ``Gans trained by a two time-scale update rule converge to a local nash equilibrium,'' \emph{Advances in neural information processing systems}, vol.~30, 2017.

\bibitem{shaham2019singan}
T.~R. Shaham, T.~Dekel, and T.~Michaeli, ``Singan: Learning a generative model from a single natural image,'' in \emph{Proceedings of the IEEE/CVF International Conference on Computer Vision}, 2019, pp. 4570--4580.

\bibitem{wright2022artfid}
M.~Wright and B.~Ommer, ``Artfid: Quantitative evaluation of neural style transfer,'' in \emph{Pattern Recognition: 44th DAGM German Conference, DAGM GCPR 2022, Konstanz, Germany, September 27--30, 2022, Proceedings}.\hskip 1em plus 0.5em minus 0.4em\relax Springer, 2022, pp. 560--576.

\bibitem{karras2020analyzing}
T.~Karras, S.~Laine, M.~Aittala, J.~Hellsten, J.~Lehtinen, and T.~Aila, ``Analyzing and improving the image quality of stylegan,'' in \emph{Proceedings of the IEEE/CVF conference on computer vision and pattern recognition}, 2020, pp. 8110--8119.

\bibitem{lai2018learning}
W.-S. Lai, J.-B. Huang, O.~Wang, E.~Shechtman, E.~Yumer, and M.-H. Yang, ``Learning blind video temporal consistency,'' in \emph{Proceedings of the European conference on computer vision (ECCV)}, 2018, pp. 170--185.

\bibitem{dosovitskiy2015flownet}
A.~Dosovitskiy, P.~Fischer, E.~Ilg, P.~Hausser, C.~Hazirbas, V.~Golkov, P.~Van Der~Smagt, D.~Cremers, and T.~Brox, ``Flownet: Learning optical flow with convolutional networks,'' in \emph{Proceedings of the IEEE international conference on computer vision}, 2015, pp. 2758--2766.

\bibitem{zhang2019exploiting}
H.~Zhang, C.~Shen, Y.~Li, Y.~Cao, Y.~Liu, and Y.~Yan, ``Exploiting temporal consistency for real-time video depth estimation,'' in \emph{Proceedings of the IEEE/CVF International Conference on Computer Vision}, 2019, pp. 1725--1734.

\bibitem{meyes2019ablation}
R.~Meyes, M.~Lu, C.~W. de~Puiseau, and T.~Meisen, ``Ablation studies in artificial neural networks,'' \emph{arXiv preprint arXiv:1901.08644}, 2019.

\bibitem{man2022review}
K.~Man and J.~Chahl, ``A review of synthetic image data and its use in computer vision,'' \emph{Journal of Imaging}, vol.~8, no.~11, p. 310, 2022.

\bibitem{tate2023}
\BIBentryALTinterwordspacing
Tate, ``Art terms,'' 2023. [Online]. Available: \url{https://www.tate.org.uk/art/art-terms/}
\BIBentrySTDinterwordspacing

\bibitem{acm2023}
\BIBentryALTinterwordspacing
ACM, ``{ACM Transactions on Graphics Author Guidelines: ACM Digital Library},'' 2023. [Online]. Available: \url{https://dl.acm.org/journal/tog/author-guidelines}
\BIBentrySTDinterwordspacing

\bibitem{malpica2023tutorial}
S.~Malpica, Q.~Sun, P.~Kellnhofer, A.~Beacco, G.~Senel, R.~McDonnell, and M.~Flores~Vargas, ``{Effective User Studies in Computer Graphics},'' in \emph{Eurographics 2023 - Tutorials}, A.~Serrano and P.~Slusallek, Eds.\hskip 1em plus 0.5em minus 0.4em\relax The Eurographics Association, 2023.

\bibitem{cowley2022framework}
H.~P. Cowley, M.~Natter, K.~Gray-Roncal, R.~E. Rhodes, E.~C. Johnson, N.~Drenkow, T.~M. Shead, F.~S. Chance, B.~Wester, and W.~Gray-Roncal, ``A framework for rigorous evaluation of human performance in human and machine learning comparison studies,'' \emph{Scientific Reports}, vol.~12, no.~1, p. 5444, 2022.

\bibitem{so2023measuring}
C.~So, ``Measuring aesthetic preferences of neural style transfer: More precision with the two-alternative-forced-choice task,'' \emph{International Journal of Human--Computer Interaction}, vol.~39, no.~4, pp. 755--775, 2023.

\bibitem{lykken1968statistical}
D.~T. Lykken, ``Statistical significance in psychological research.'' \emph{Psychological bulletin}, vol.~70, no. 3, Pt.1, pp. 151--159, 1968.

\bibitem{ellis2010essential}
P.~D. Ellis, \emph{The essential guide to effect sizes: Statistical power, meta-analysis, and the interpretation of research results}.\hskip 1em plus 0.5em minus 0.4em\relax Cambridge university press, 2010.

\bibitem{duca2022}
\BIBentryALTinterwordspacing
A.~L. Duca, ``Hypothesis testing explained,'' May 2022. [Online]. Available: \url{https://www.kdnuggets.com/2021/09/hypothesis-testing-explained.html}
\BIBentrySTDinterwordspacing

\bibitem{friedman1937use}
M.~Friedman, ``The use of ranks to avoid the assumption of normality implicit in the analysis of variance,'' \emph{Journal of the american statistical association}, vol.~32, no. 200, pp. 675--701, 1937.

\bibitem{wilcoxon1992individual}
F.~Wilcoxon, ``Individual comparisons by ranking methods,'' in \emph{Breakthroughs in Statistics: Methodology and Distribution}.\hskip 1em plus 0.5em minus 0.4em\relax Springer, 1992, pp. 196--202.

\bibitem{cohen2013statistical}
J.~Cohen, \emph{Statistical power analysis for the behavioral sciences}.\hskip 1em plus 0.5em minus 0.4em\relax Academic press, 2013.

\bibitem{cliff1993dominance}
N.~Cliff, ``Dominance statistics: Ordinal analyses to answer ordinal questions.'' \emph{Psychological bulletin}, vol. 114, no.~3, p. 494, 1993.

\bibitem{chen2023collaborative}
H.~Chen, F.~Shao, X.~Chai, Q.~Jiang, X.~Meng, and Y.-S. Ho, ``Collaborative learning and style-adaptive pooling network for perceptual evaluation of arbitrary style transfer,'' \emph{IEEE Transactions on Neural Networks and Learning Systems}, 2023.

\bibitem{murray2012ava}
N.~Murray, L.~Marchesotti, and F.~Perronnin, ``{AVA}: A large-scale database for aesthetic visual analysis,'' in \emph{2012 IEEE conference on computer vision and pattern recognition}.\hskip 1em plus 0.5em minus 0.4em\relax IEEE, 2012, pp. 2408--2415.

\bibitem{achlioptas2021artemis}
P.~Achlioptas, M.~Ovsjanikov, K.~Haydarov, M.~Elhoseiny, and L.~J. Guibas, ``Artemis: Affective language for visual art,'' in \emph{Proceedings of the IEEE/CVF Conference on Computer Vision and Pattern Recognition}, 2021, pp. 11\,569--11\,579.

\bibitem{talebi2018nima}
H.~Talebi and P.~Milanfar, ``{NIMA}: Neural image assessment,'' \emph{IEEE transactions on image processing}, vol.~27, no.~8, pp. 3998--4011, 2018.

\bibitem{yi2023towards}
R.~Yi, H.~Tian, Z.~Gu, Y.-K. Lai, and P.~L. Rosin, ``Towards artistic image aesthetics assessment: a large-scale dataset and a new method,'' in \emph{Proceedings of the IEEE/CVF Conference on Computer Vision and Pattern Recognition}, 2023, pp. 22\,388--22\,397.

\end{thebibliography}

\newpage

 




\clearpage

{\appendices

\section*{Evaluation Data}
Figure~\ref{fig:styles-content-common_2} provides examples of content and style images commonly utilized for the evaluation of NST methods.

\begin{figure}[htb]
    \centering
    \includegraphics[width=\linewidth]{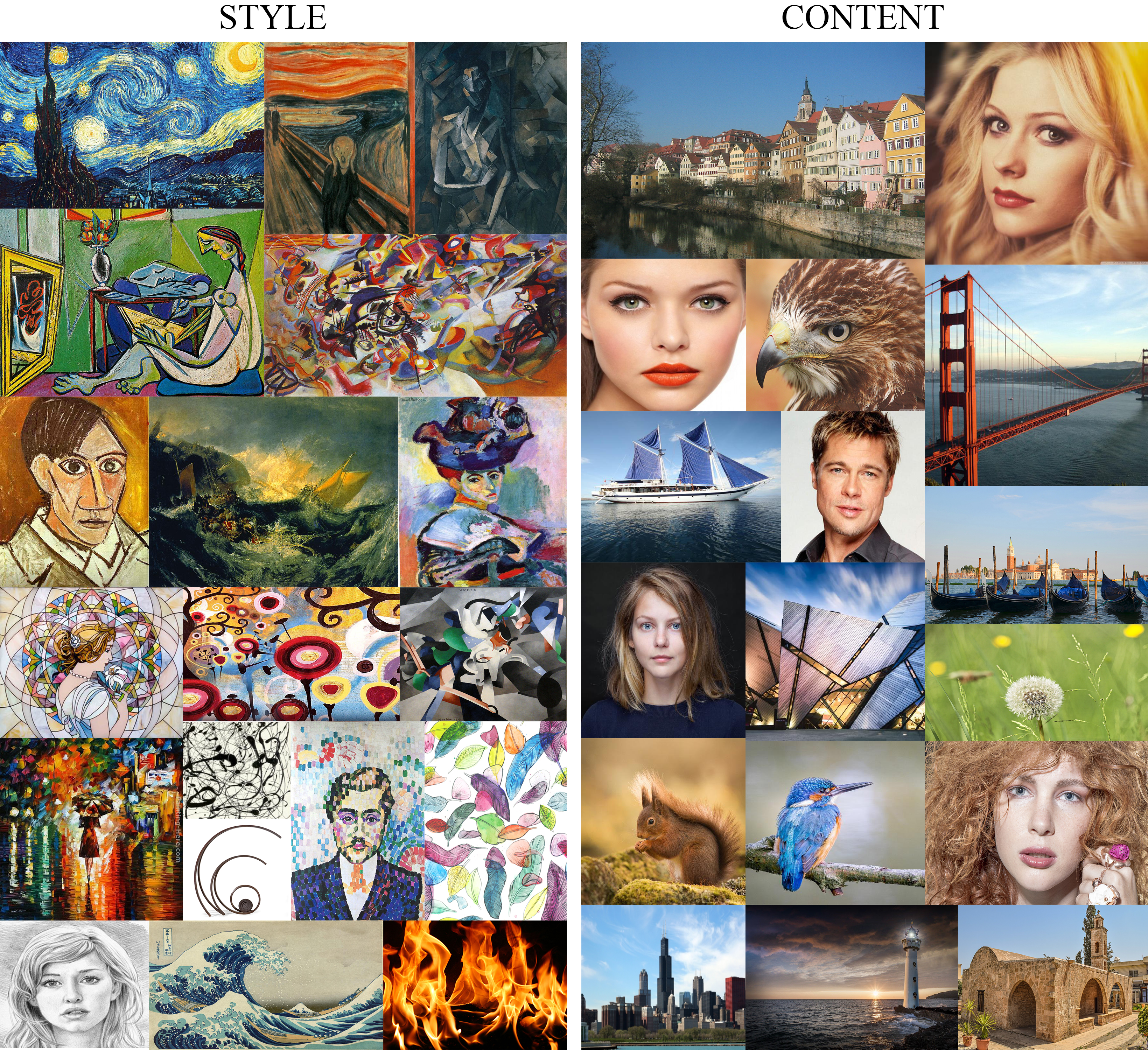}
    \caption[Most used content and style images for Qualitative Evaluation]{The most common content and style images used in the Qualitative Evaluation sections of NST studies. Almost all the studies include at least some of the images displayed here.}
     \label{fig:styles-content-common_2}
\end{figure}

\section*{Experiments}
Figure~\ref{fig:experiments_2} demonstrates more qualitative results from state-of-the-art methods.

\begin{figure*}[hb]
    \centering
    \includegraphics[width=\linewidth]{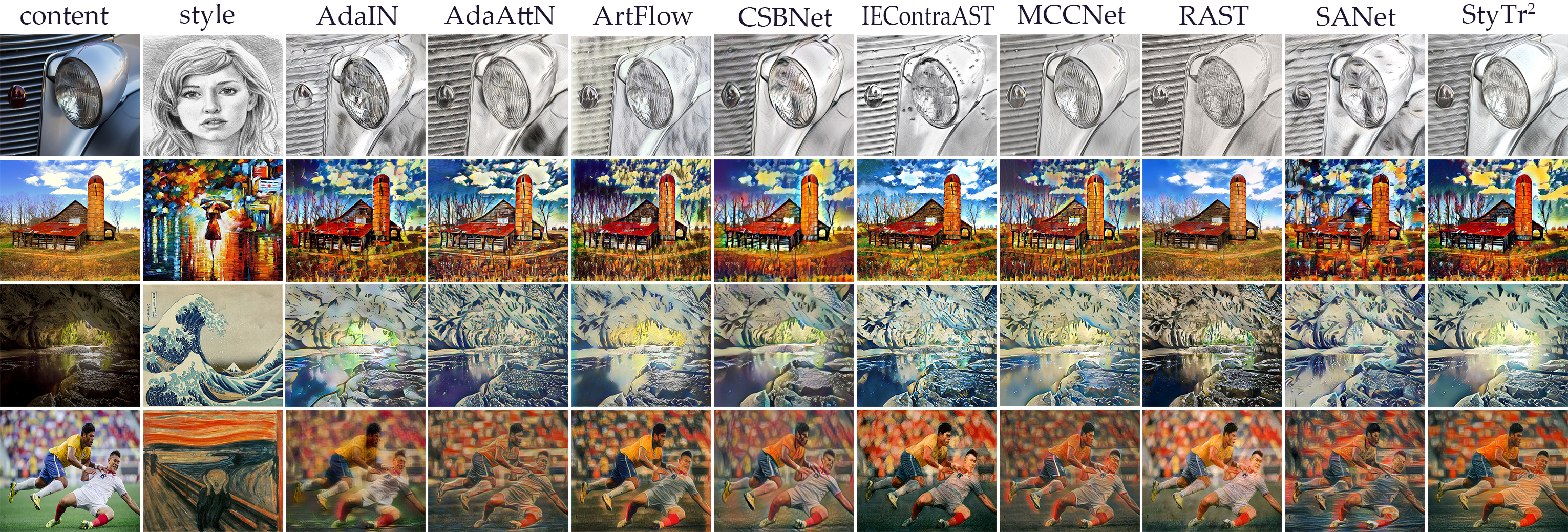}
    \caption[Results of state-of-the-art NST methods]{Results of state-of-the-art NST approaches.}
     \label{fig:experiments_2}
\end{figure*}

}

\vfill

\end{document}